\documentclass[10pt,twocolumn,letterpaper]{article}

\usepackage[pagenumbers]{cvpr} 

\usepackage{graphicx}
\usepackage{amsmath}
\usepackage{amssymb}
\usepackage{booktabs}
\usepackage{dsfont}
\usepackage{multirow} 
\usepackage{multicol}
\usepackage{adjustbox}
\usepackage{pifont}

\usepackage{algorithm}
\usepackage{algorithmic}
\usepackage{verbatim}
\newcommand*\samethanks[1][\value{footnote}]{\footnotemark[#1]}

\usepackage[pagebackref,breaklinks,colorlinks]{hyperref}

\usepackage[capitalize]{cleveref}
\crefname{section}{Sec.}{Secs.}
\Crefname{section}{Section}{Sections}
\Crefname{table}{Table}{Tables}
\crefname{table}{Tab.}{Tabs.}

\def\cvprPaperID{2053} 
\def\confName{CVPR}
\def\confYear{2023}

\begin{document}

\title{Continual Semantic Segmentation with Automatic Memory Sample Selection}

\author{
    Lanyun Zhu\textsuperscript{\rm1}\thanks{Equal Contribution} ~~~
    Tianrun Chen\textsuperscript{\rm2}\samethanks[1] ~~~
    Jianxiong Yin\textsuperscript{\rm3} ~~~
    Simon See\textsuperscript{\rm3} ~~~ 
    Jun Liu\textsuperscript{\rm1}\thanks{Corresponding Author} ~~~ \\
    Singapore University of Technology and Design \textsuperscript{\rm1} ~~~
    Zhejiang University \textsuperscript{\rm2} ~~~
    NVIDIA AI Tech Centre \textsuperscript{\rm3}\\
    {\tt\small lanyun\_zhu@mymail.sutd.edu.sg} ~~~
    {\tt\small tianrun.chen@zju.edu.cn} \\
    {\tt\small \{jianxiongy, ssee\}@nvidia.com} ~~~
    {\tt\small jun\_liu@sutd.edu.sg}\\
}
\maketitle

\begin{abstract}
Continual Semantic Segmentation (CSS) extends static semantic segmentation by incrementally introducing new classes for training. To alleviate the catastrophic forgetting issue in CSS, a memory buffer that stores a small number of samples from the previous classes is constructed for replay. However, existing methods select the memory samples either randomly or based on a single-factor-driven hand-crafted strategy, which has no guarantee to be optimal. In this work, we propose a novel memory sample selection mechanism that selects informative samples for effective replay in a fully automatic way by considering comprehensive factors including sample diversity and class performance. Our mechanism regards the selection operation as a decision-making process and learns an optimal selection policy that directly maximizes the validation performance on a reward set. To facilitate the selection decision, we design a novel state representation and a dual-stage action space. Our extensive experiments on Pascal-VOC 2012 and ADE 20K datasets demonstrate the effectiveness of our approach with state-of-the-art (SOTA) performance achieved, outperforming the second-place one by 12.54\% for the 6-stage setting on Pascal-VOC 2012.
\end{abstract}

\section{Introduction}
Semantic segmentation is an important task with a lot of applications. The rapid development of algorithms \cite{deeplabv3, sfnet, zhu2021learning, Gu_2022_CVPR, fu2022panoptic, liu2021label} and the growing number of publicly available large datasets \cite{Cordts2016Cityscapes, zhou2017scene} have led to great success in the field. 
However, in many scenarios, the static model cannot always meet real-world demands, as the constantly changing environment calls for the model to be constantly updated to deal with new data, sometimes with new classes. 

A naive solution is to apply continual learning by incrementally adding new classes to train the model. However, it is not simple as it looks -- almost every time, since the previous classes are inaccessible in the new stage, the model forgets the information of them after training for the new classes. This phenomenon, namely catastrophic forgetting, has been a long-standing issue in the field. Furthermore, the issue is especially severe in dense prediction tasks like semantic segmentation. 

Facing the issue, existing works \cite{tiwari2022gcr, rebuffi2017icarl, borsos2020coresets, isele2018selective, aljundi2019gradient, jin2020gradient, bang2021rainbow, douillard2021plop, cha2021ssul} propose to perform exemplar replay by introducing a memory buffer to store some samples from previous classes. By doing so, the model can be trained with samples from both current and previous classes, resulting in better generalization. However, since the number of selected samples in the memory is much smaller than those within the new classes, the selected samples are easy to be ignored or cause overfitting when training due to the small number. Careful selection of the samples is required, which naturally brings the question: \textit{How to select the best samples for replay?}

Some attempts have been made to answer the question, aiming to seek the most effective samples for replay. Researchers propose different criteria that are mostly manually designed based on some heuristic factors like diversity \cite{tiwari2022gcr, rebuffi2017icarl, borsos2020coresets, isele2018selective, aljundi2019gradient, jin2020gradient, bang2021rainbow}. For example, \cite{michieli2019incremental} selects the most common samples with the lowest diversity for replay, believing that the most representative samples will elevate the effectiveness of replay. However, the most common samples may not always be the samples being forgotten in later stages.   
\cite{bang2021rainbow} proposes to save both the low-diversity samples near the distribution center and high-diversity samples near the classification boundaries. However, new challenges arise since the memory length is limited, so it is challenging to find the optimal quotas for the two kinds of samples to promote replay effectiveness to the greatest extent. Moreover, most of the existing methods are designed based on a single factor, the selection performance, however, can be influenced by many factors with complicated relationships. For example, besides diversity, memory sample selection should also be \textit{class-dependent} because the hard classes need more samples to replay in order to alleviate the more severe catastrophic forgetting issue. 
Therefore, we argue that it is necessary to select memory samples in a more intelligent way by considering the more comprehensive factors and their complicated relationships.

Witnessing the challenge, in this work, we propose a novel automatic sample selection mechanism for CSS. Our key insight is that selecting memory samples can be regarded as a decision-making task in different training stages, so we formulate the sample selection process as a Markov Decision Process, and we propose to solve it automatically with a reinforcement learning (RL) framework. Specifically, we employ an agent network to make the selection decision, which receives the state representation as the input and selects optimal samples for replay. To help the agent make wiser decisions, we construct a novel and comprehensive state combined with the sample diversity and class performance features. In the process of state computation, the inter-sample similarity needs to be measured. We found the naive similarity measurement by computing the prototype distance is ineffective in segmentation, as the prototype losses the local structure details that are important for making pixel-level predictions. Therefore, we propose a novel similarity measured in a multi-structure graph space to get a more informative state. We further propose a dual-stage action space, in which the agent not only selects the most appropriate samples to update the memory, but also enhances the selected samples to have better replay effectiveness in a gradient manner. All the careful designs allow the RL mechanism to be effective in solving the sample selection problem for CSS.   

We perform extensive experiments on Pascal-VOC 2012 and ADE 20K datasets, which demonstrate the effectiveness of our proposed novel paradigm for CSS. Benefiting from the reward-driven optimization, the automatically learned policy can help select the more effective samples, thus resulting in better performance than the previous strategies.  On both datasets, our method achieves state-of-the-art (SOTA) performance. To summarize, our contributions are as follows:

\begin{itemize}
\item We formulate the sample selection of CSS as a Markov Decision Process, and introduce a novel and effective automatic paradigm for sample replay in CSS enabled by reinforcement learning. 
\item We design an effective RL paradigm tailored for CSS, with novel state representations containing multiple factors that can guide the selection decision, and a dual-stage action space to select samples and boost their replay effectiveness.
\item  Extensive experiments demonstrate our automatic paradigm for sample replay can effectively alleviate the catastrophic forgetting issue with state-of-the-art (SOTA) performance achieved. 
\end{itemize}
\begin{figure*}[t]
	\centering
	
	\includegraphics[width=\linewidth]{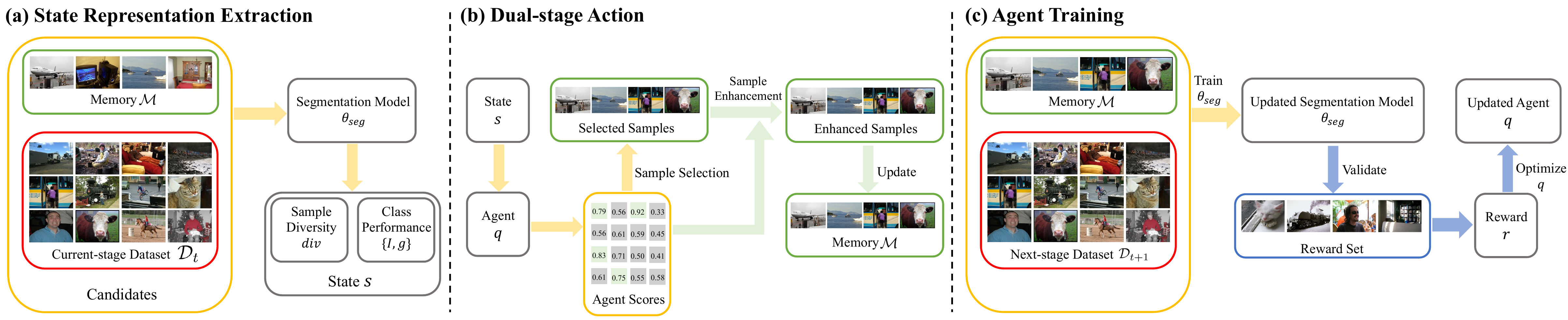}
	
	\caption{\textbf{The overall scheme of our automatic memory sample selection mechanism for CSS.} (a) Given the memory $\mathcal{M}$ and current-stage dataset $\mathcal{D}_{t}$, we first extract the state representation for each sample in $\mathcal{M}\cup \mathcal{D}_{t}$, which is consisted of the sample diversity and class performance features. (b) Given the state representations, the agent $q$ produces a score for each candidate sample. Based on the scores, we select several samples and enhance them in a gradient-based manner. The memory $\mathcal{M}$ is updated by these samples. (c) The segmentation model $\theta_{seg}$ is trained using the updated $\mathcal{M}$ and $\mathcal{D}_{t+1}$. We then validate the updated $\theta_{seg}$ on a reward set, resulting in the reward $t$ that is used to optimize agent $q$.
	}
		\label{overall}
\end{figure*}
\vspace{-0.9\baselineskip}
\section{Related Work}
\noindent \textbf{Semantic Segmentation and Continual Semantic Segmentation.} Semantic segmentation is a basic task in computer vision and has achieved great success in recent years benefiting from the rapid development of deep-learned based algorithms such as encoder-decoder structure \cite{unet, bisenet, bisenetv2, segnet, sfnet}, dilated convolution \cite{deeplabv1, deeplabv2, deeplabv3, deeplabv3+}, pyramid structure \cite{zhu2021learning, deeplabv3, pspnet, deeplabv3+}, attention mechanism \cite{acfnet, danet, ann} and transformers \cite{zheng2021rethinking, xie2021segformer, strudel2021segmenter, cheng2022masked}. To meet the requirement in real applications where the new classes are incrementally added, continual learning has been proposed \cite{chaudhry2018riemannian, dhar2019learning, pourkeshavarzi2021looking, qin2021bns} and applied to the semantic segmentation task \cite{michieli2019incremental, cermelli2020modeling, zhang2022representation, phan2022class, michieli2021continual, douillard2021plop}. Among them, many works adopt replay-based methods, which show high effectiveness. \cite{cha2021ssul, yan2021framework} use a memory buffer to store replay exemplars, however, in which the samples are selected either randomly or according to heuristic rules. \cite{douillard2021plop} derives richer replay exemplars through a generative adversarial network with high computation cost or web-crawled images requiring the extra data. Different from the above methods, with an RL-driven automatic memory selection policy and the gradient-based sample enhancement operation, our method can be very effective for CSS. \\

\noindent \textbf{Memory Sample Selection.} How to select the appropriate samples is a severe issue for replay-based continual learning methods. Most the previous selection methods rely on manually-designed strategies based on heuristic rules such as sample diversity \cite{aljundi2019gradient, yoon2021online, jin2020gradient, bang2021rainbow}, adversarial Shapley value \cite{shim2021online} or feature matching \cite{rebuffi2017icarl}. In general, such hand-crafted methods lack effectiveness guarantees and are difficult to be optimal due to a complex interplay between factors that affect selection performance, as discussed in the Introduction. Our method explores a novel direction by enabling the selection policy to be automatically learned with a carefully-designed RL mechanism.     \\

\noindent \textbf{Reinforcement Learning. }Reinforcement learning (RL) has achieved remarkable success in many decision-making tasks like game intelligence \cite{silver2016mastering} and robot control \cite{johannink2019residual, ibarz2021train}. It has also been employed to computer vision with various applications such as active learning \cite{gong2022meta}, pose estimation \cite{guo2022visual}, model compression \cite{alwani2022decore} and person re-identification \cite{Wu_2022_CVPR}. \cite{liu2021rmm} uses RL for the exemplars length management, however, with the completely different working mechanism from ours. Instead of employing RL to control class-level memory length and then still needing a random selection process, our method is end-to-end and can directly select specific samples in one step fully automatically, showing significant effectiveness in semantic segmentation with the task-tailored state representations and a novel dual-stage action space.

\section{Preliminaries}
Continual semantic segmentation (CSS) aims to train a segmentation model in $T$ stages continuously without forgetting. In each stage $t$, a training dataset $\mathcal{D}_{t}$ can be utilized, where only pixels within the current classes $\mathcal{C}_{t}$ are labeled, leaving pixels within others classes (including previous classes $\mathcal{C}_{1:t-1}$ and future classes $\mathcal{C}_{t+1:T}$) as the background class. The goal is to allow the model to be able to predict all classes $\mathcal{C}_{1:T}$ after completing all $T$ stages. To alleviate the catastrophic forgetting problem in CSS, an exemplar memory $\mathcal{M}$ that contains a small number of sampled data from the previous classes can be  used for replay, so that both $\mathcal{M}$ and $\mathcal{D}_{t}$ are involved for training. 


In the training process, $\mathcal{M}$ is updated once a training stage is completed. This means $\mathcal{M}$ will be refilled by new samples from $\mathcal{M} \cup \mathcal{D}_{t}$ after the stage $t$ with the learning on $\mathcal{D}_{t}$ completed. It is obvious that the careful selection of samples for $\mathcal{M}$ could greatly affect the performance, which is also the focus of this work. 

\section{Method} \label{selection}
\subsection{Overall}
Considering the memory $\mathcal{M}$ with $L$ samples and $\mathcal{D}_{t}$ with $N_{t}$ samples, the target of this work is to learn an optimal policy that automatically selects $L$ samples from $\mathcal{M} \cup \mathcal{D}_{t}$ and put them into $\mathcal{M}$ for the next stage training, driven by maximizing the designed reward reflecting the performance improvement. 
The selection decision is made by an agent network that is a three-layered MLP. It converts the CSS to become a decision-making process with the following 
procedure: 
    1) Obtaining the state $s$ by assessing the properties of samples that can measure its contribution for replay. 2) Based on $s$, using the agent $q$ to make an action $a$ that selects $L$ samples to update the memory $\mathcal{M}$. 
    3) Training the segmentation network with the updated $\mathcal{M}$. 4) Computing the reward $r$ based on the validation performance of the updated segmentation network. 5) Repeating the above steps until completing all $T$ stages. 6) Optimizing agent $q$ based on $r$ from all stages. 

As shown in Fig.\ref{overall}, in this work, we solve the above problem under a reinforcement learning (RL) framework, in which the agent $q$ scores each state $s$ and makes an action $a$ based on the score. Benefiting from the task-specific state representations, a novel selection-enhancement dual-stage action space and the reward-driven optimization, we can optimize the agent to learn an effective selection policy. In the following parts of this section, we illustrate the details of how these components are designed. 

\subsection{State Representation} \label{state}
\vspace{-0.7\baselineskip}
The state representation $s$ is the key to making the automatic selection decision process possible, as it is the input to and serves as the decision support of the agent network. Designing the state should consider the requirements of the selection policy. Intuitively, an optimal policy should make a selection decision by estimating the potential replay contribution of each sample, and allocate different quotas to different classes as the hard classes suffer from the more severe catastrophic forgetting issue and need more samples to replay. Based on these intuitions, we propose to combine two kinds of cues including \textit{\textbf{sample diversity}} and \textbf{\textit{class performance}} for constructing state. For an image within class $c$, sample diversity $div$ measures its novelty, which can reflect the potential replay effectiveness as indicated by previous works \cite{bang2021rainbow, rebuffi2017icarl}. A higher $div$ indicates the sample differs more from other images within the same class $c$. We calculate it by computing and averaging the inter-sample similarities. The class performance is constructed as the combination of two metrics: 1) accuracy and 2) forgetfulness. We derive accuracy by computing the training IoU $I_{c}$ for each class $c$. The hard classes that are trained to the worse performance have the lower IoUs. However, as the IoU measures the current training accuracy, it cannot reflect whether a class is easily forgotten in the future, which is critical for CSS but difficult to measure directly since the future performance is unknown. We thus estimate forgetfulness $g_{c}$ by measuring the similarities between $c$ with all other classes, motivated by the previous finding that classes that are more similar to other classes are more likely to be forgotten \cite{phan2022class}. 
Eventually, given an image, on all $C$ classes in it, we compute their diversities $\{{div}_{c}\}_{c=1}^{C}$, accuracy $\{I_{c}\}_{c=1}^{C}$ and forgetfulness $\{g_{c}\}_{c=1}^{C}$, resulting in three groups of features. Then, we calculate the average values of the three groups over different classes, and concatenate them to get the state representation $s$ of the image.

\subsubsection{Measuring Similarity in Multi-structure Space}
\noindent \textbf{Motivation.} Both the sample diversity $div$ and forgetfulness $g_{c}$ introduced above need to compute the similarity. 
In previous works, the similarity is mainly measured in the \textit{prototype-level space} \cite{rebuffi2017icarl} or \textit{pixel-level space} \cite{wang2020few}. The former condenses the sample into a single prototype feature and then calculates the feature distance. It is computationally efficient, but drops the spatial information and structural details, which leads to errors. For example, two images with completely different local structures or object postures may have similar prototype features, since the prototypes are computed by the average features of all pixels, concealing the differences between local details. Such errors caused by the lack of local details are detrimental to the segmentation task, where local structural information is important for making pixel-level predictions \cite{zhu2021learning}. As a result, the state constructed by the prototype-level similarity leads to poor performance when employed to CSS. The pixel-level one retains the local information, however, it requires an unacceptable computation cost due to the pixel-wise distance calculation and may cause overfitting \cite{li2021adaptive}. Thus, to obtain a more informative similarity, a novel representation space is needed, which should not only retain the spatial and structural information but also be condensed for a reasonable computation cost. Based on the discussion, we propose a novel method that first maps each sample into a \textit{multi-structure graph space} and then measures the inter-sample similarity based on the graph matching. Each vertex of the graph represents a semantic structure, and the edge represents the spatial and semantic correlations, thus a fine-grained similarity can be measured by utilizing the comprehensive information.\\

\begin{figure}
    \centering
    \includegraphics[width=0.9\linewidth]{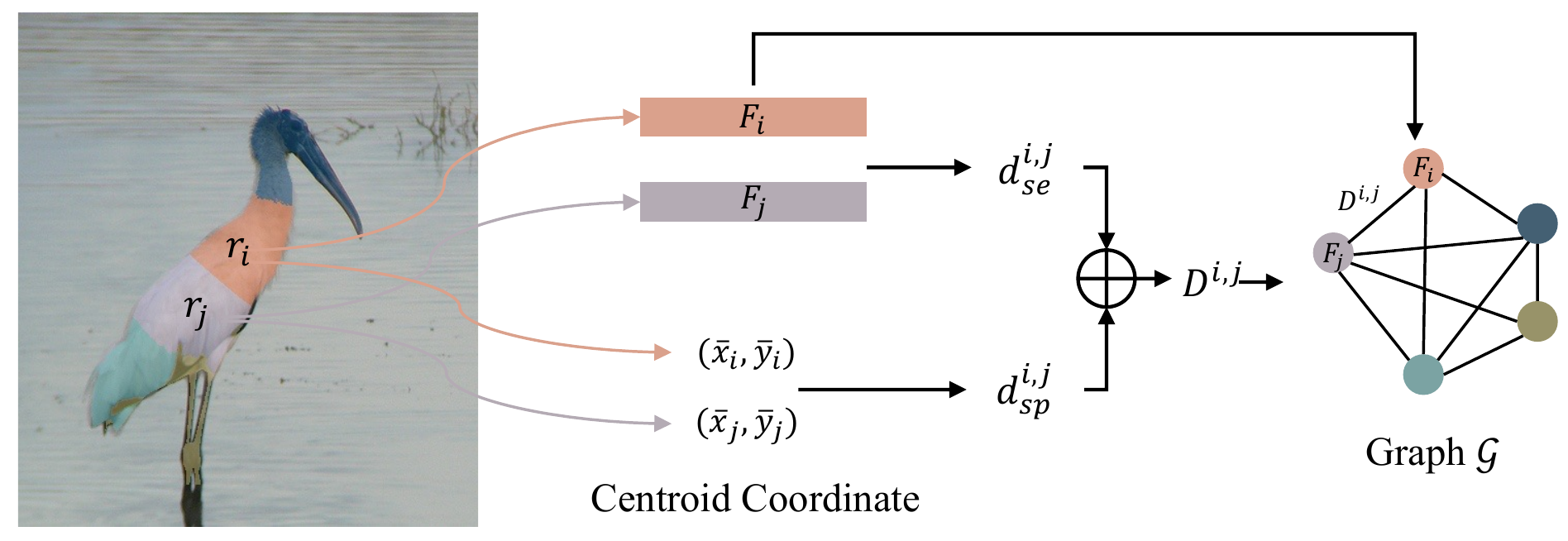}
    \caption{Illustration of how the graph for computing sample diversity is constructed. In the figure, $r_{i}$ and $r_{j}$ denote two superpixels. $F_{i}$ and $F_{j}$ refer to the average features for all pixels within them. $(\overline{x}_{i}, \overline{y}_{i})$ and $(\overline{x}_{j}, \overline{y}_{j})$ denote the centroid coordinates of $r_{i}$ and $r_{j}$ respectively. $d_{se}^{i,j}$ and $d_{sp}^{i,j}$ refer to the semantic distance and spatial distance. The generated graph $\mathcal{G}$ will be used to compute the sample diversity.}
    \label{graph_fig}
\end{figure}

\noindent \textbf{Multi-structure Graph.}
Considering an image with the class $c$, we represent the region $\mathcal{R}$ within $c$ as a graph $\mathcal{G}$ through the way illustrated by Fig. \ref{graph_fig}. To get the local structural representation, we first use the method as in \cite{li2021adaptive} to generate $M$ superpixels $\{r_m{}\}_{m=1}^{M}$ ($r_{1} \cup r_{2} \cup ... \cup r_{M}= \mathcal{R}$). The motivation for using superpixels is that, according to the construction mechanism of superpixels, each $r_{m}$ can represent a meaningful semantic structure such as the head of a bird, and condenses the pixel-level representation enabled by clustering pixels with similar features and adjacent positions.
Each vertex $F_{m}$ is then computed as the average feature for all pixels within $r_{m}$. We represent the edge of $\mathcal{G}$ as a distance map $D\in \mathbb{R}^{M\times M}$, where the element $D^{i, j}$ denotes the distance between the $i$-th and $j$-th vertices. 
To simultaneously consider the context-aware high-level semantic information and low-level spatial correlation, we combine both the \textit{semantic distance} and \textit{spatial distance} for getting $D$. Concretely, the semantic distance $d_{se}^{i, j}$ is the L2 distance between $F_{i}$ and $F_{j}$; the spatial distance $d_{sp}^{i, j}$ denotes the Euclidean distance between the two centroid coordinates \footnote{Considering a superpixel $r = \{(x_{i},y_{i})\}_{i=1}^{N}$, the centroid coordinate $(\overline{x}, \overline{y})$ is computed as: $\overline{x} = \frac{1}{N}\sum_{i=1}^{N}x_{i}$,\; $\overline{y} = \frac{1}{N}\sum_{i=1}^{N}y_{i}$.} of the superpixels $r_{i}$ and $r_{j}$, reflecting their relative positions. We normalize $d_{se}^{i,j}$ and $d_{sp}^{i,j}$ to $[0, 1]$ and derive $D^{i,j} = d_{se}^{i,j} + d_{sp}^{i,j}$. Such a graph can capture comprehensive representations such as local structure details and spatial information, which are lost in the prototype space but are crucial for measuring a fine-grained similarity.\\ 

\noindent \textbf{Inter-graph Similarity.}
After mapping images into the graph space, we use the matching algorithm to measure the similarities. For two graphs $\mathcal{G}_{i}$ and $\mathcal{G}_{j}$, the Sinkhorn algorithm \cite{cuturi2013sinkhorn} is applied for aligning them, in which the transport cost $tc$ is obtained by solving the optimal transport problem. A higher $tc$ represents the lower similarity of the two graphs. The details for this step are presented in supplementary materials. As the edge distance $D^{i,j}$ is computed with both the semantic and spatial distance, the computed $tc$ after matching can reflect both the semantic and spatial similarity. For example, considering two regions for the `person' class, we can measure both whether they wear similar clothes (semantic similarity) and whether they are with the same body posture (spatial similarity), capturing the comprehensive fine-grained representations.\\ 

\noindent \textbf{Representation Computation.}
We use the above-mentioned similarity measurement to compute the sample diversity $div$ and forgetfulness $g$ in state representations. For an image with the $c$-th class, let $\mathcal{G}$ be its graph. We introduce a support set $\mathcal{S}_{c} = \{\mathcal{G}_{c}^{i}\}_{i=1}^{N_{c}}$ to contain several graphs for other images within the same class $c$. For each previous class in $\mathcal{C}_{1:t-1}$, we construct $\mathcal{S}_{c}$ as the set of all images saved in the memory. For each current class in $\mathcal{C}_{t}$ that has a larger number of samples, to relieve the computation burden, we randomly sample 10\% from all images to form $\mathcal{S}_{c}$. We will show in supplementary material that $div$ computed from a sampled set can be effective enough. A diverse and novel sample is likely to have low similarities compared to other samples within the same class. We thus get $div$ by computing the average similarities by:
\begin{equation}
    div = \frac{1}{|\mathcal{S}_{c}|}\sum\limits_{\mathcal{G}_{c}^{i}\in \mathcal{S}_{c}}{\rm Sim}\left(\mathcal{G}, \mathcal{G}_{c}^{i}\right),
\end{equation}
where ${\rm Sim}$ refers to the inter-graph similarity measurement introduced above. To get the forgetfulness $g_{c}$ for each class $c$, we first construct a representative set $\hat{\mathcal{S}}_{c} = \{\mathcal{G}_{c}^{i}\}_{i=1}^{\hat{N}_{c}}$ containing the top 10\% samples in $\mathcal{S}_{c}$ with the lowest diversity scores. These samples are most similar to other samples in $c$ so they can represent the class-level properties. Then forgetfulness $g_{c}$ is gotten as the class-wise similarity computed by:
\begin{equation}
    g_{c} = \frac{1}{|\hat{\mathcal{S}}_{c}|}\sum\limits_{\mathcal{G}_{c}^{i} \in \hat{\mathcal{S}}_{c}}\frac{1}{|\mathcal{C}_{1:t}|-1}\sum\limits_{j\in \mathcal{C}_{1:t}\backslash c}\frac{1}{|\hat{\mathcal{S}}_{j}|}\sum \limits_{\mathcal{G}_{j}^{k}\in \hat{\mathcal{S}}_{j}}{\rm Sim }\left(\mathcal{G}_{c}^{i}, \mathcal{G}_{j}^{k}\right).
\end{equation}
 Eventually, the obtained $div$ and $g$ are combined with the accuracy $I$, generating the state representations that can help make a wiser selection decision.

\subsection{Dual-stage Action with Sample Selection and Enhancement}
\label{selection_enhancement}
After getting the state information $s^{i}$ for each sample, we use an agent network $q$ to produce a score $q(s^{i})$ by taking $s^{i}$ as the input. A higher
score indicates the sample is more suitable for replay. Thus, we regard agent score as the replay effectiveness
indicator, and utilize it to drive a novel action space for the RL mechanism that has two stages: \textit{sample selection} and \textit{sample enhancement}. 


Concretely, we first select memory samples by $L$ ones with the highest agent scores, which is written as:
\begin{equation}
    a = \mathop{{\rm TopL}}\limits_{i\in [1,L+N_{t}]}q\left({s}^{i}\right).
\end{equation}

After that, instead of directly using the static selected samples for training in the next stage, we
further propose an enhancement operation that
edits each sample to be more effective for replay. This is motivated by our observation of the agent scores for the selected samples. We notice that, only 10\%
of the selected samples have agent scores exceeding
0.8 (the theoretical maximum score is 1). The phenomenon shows that such samples are the best possible choice from the imperfect candidates, but not the ideally perfect samples for replay. Thus, despite achieving better performance by selecting the most adequate samples, there is still room to further improve the replay effectiveness if we can enhance the samples to reach higher scores. We thus implement enhancement through a gradient-based manner by maximizing the agent score. Concretely, we regard the state $s^{x}$ as a feature computed from input image $x$ along with $\mathcal{M}$ and $\mathcal{D}_{t}$ under the segmentation network parameters $\theta_{seg}$ with the state computing function $f_{s}$, which is formulated as:
\begin{equation}
    s^{x} = f_{s}\left(x; \mathcal{M}, \mathcal{D}_{t}, \theta_{seg}\right).
\end{equation}
Then the agent score is generated by $q(s^{x})$. We perform a gradient update on $x$ so that the agent  score $q(s^{x})$ moves towards the larger direction reflecting the better replay effectiveness, which is written as:
\begin{equation} \label{gradient}
\begin{aligned}
    x' &= x + \epsilon \nabla_{x}q\left(s^{x}\right) \\&=  x + \epsilon \nabla_{x}q\left(f_{s}\left(x; \mathcal{M}, \mathcal{D}_{t}, \theta_{seg}\right)\right) ,
\end{aligned}
\end{equation}
where $\epsilon$ is a hyper-parameter to control an adequate updating rate so that the image label remains unchanged. With the higher agent score, the resulted $x'$ can be more effective and is stored into $\mathcal{M}$ for replay.  

\subsection{Reward and Optimization} \label{reward_optimization}
Our selection policy aims to allow the segmentation model trained with the memory $\mathcal{M}$ to achieve better performance. Therefore, the reward for optimizing agent should reflect how much the memory samples derived by the agent policy can benefit the CSS training. To implement the goal, we divide a subset from the training set to get a reward set $\mathcal{D}^{reward}$, and define reward $r_{t}$ at the $t$-th stage as the validation accuracy on $\mathcal{D}^{reward}$ evaluated on the segmentation model that has completed the $t$-th stage. With reward derived, following DQN algorithm \cite{van2016deep}, the agent is optimized by the temporal difference (TD) error formulated as: 
\begin{equation}
\label{td_loss}
\begin{aligned}
    TD\left(\theta, \hat{\theta}\right) = \frac{1}{T-1}\sum_{t=1}^{T-1}\left(r_{t+1}+\frac{\gamma}{L}\sum_{i=1}^{L}q\left(s_{t+1}^{ a^{i}_{t+1}}; \hat{\theta}\right) \right.\\\left.- \frac{1}{L}\sum_{i=1}^{L}q\left(s_{t}^{a^{i}_{t}}; \theta \right)\right)^{2},
\end{aligned}
\end{equation}
where $s_{t}^{a^{i}_{t}}$ refers to the state representation of the $i$-th selected sample in the $t$-th stage, $\theta$ and $\hat{\theta}$ refer to the agent’s policy and off-policy parameters respectively. Following \cite{van2016deep}, $\hat{\theta}$ is periodically updated based on $\theta$, aiming to save the learned Q-value.

\begin{algorithm}[t]
\algsetup{linenosize=\scriptsize} \scriptsize
\caption{Agent Training Algorithm.}
\label{alg}
\begin{algorithmic}[1]
\STATE{\textbf{Input:} agent network $q$, segmentation network parameters $\theta_{seg}$, dataset $\mathcal{D}_{1}$. }
\FOR{$y$ \textbf{in} $1,...,Y$}
\STATE{Create a new task having $T_{y}$ continual stages with class partitions
$\{\mathcal{C}_{t_{y}}\}_{t_{y}=1}^{T_{y}}$}.
\STATE{Partition $\mathcal{D}_{1}$ to $\mathcal{D}_{1}^{train}$ and $\mathcal{D}_{1}^{reward}$}
\STATE{Initialize $\theta_{seg}$, initialize $\mathcal{M}$ as an empty set}
\FOR{$t_{y}$ \textbf{in} $1,...,T_{y}$}
\STATE{Train $\theta_{seg}$ on $\mathcal{M} \cup \mathcal{D}_{1}^{train, t_{y}}$}
\STATE{Compute state $s_{t}$ (Sec.\ref{state}) and agent scores $q(s_{t})$}
\STATE{Select and enhance samples (Sec.\ref{selection_enhancement}), update $\mathcal{M}$}
\IF{$t_{y}>1$}
\STATE{Compute reward $r_{t_{y}}$ (Sec.\ref{reward_optimization})}
\ENDIF
\ENDFOR
\STATE{Update $q$ by Eq. \ref{td_loss}}
\ENDFOR
\STATE{\textbf{Return:} $q$}
\end{algorithmic}
\end{algorithm}
\subsection{Agent Training and Deployment}
\vspace{-0.5\baselineskip}
With the above-introduced RL mechanism for CSS, we then present the agent training and deployment method in this section. We denote $\mathcal{D}_{1}$ as the dataset for first-stage training. According to CSS protocol \cite{douillard2021plop}, $\mathcal{D}_{1}$ contains multiple classes (usually more than half of the total). Thus, it can provide sufficient information for training an effective agent. The detailed training process is shown in Alg. \ref{alg}. We train the agent for $Y$ iterations. In each iteration, we randomly divide $\mathcal{D}_{1}$ into the training set $\mathcal{D}_{1}^{train}$ and the reward set $\mathcal{D}_{1}^{reward}$, and set a new CSS task by reallocating the classes observed in each stage. This helps the agent to learn a more general policy with training from diverse settings.

Once the agent training is completed, we can deploy it on the whole set $\mathcal{D} = \{\mathcal{D}_{i}\}_{i=1}^{T}$, selecting and enhancing memory samples at the end of each stage and using them for replay in the next stage.

\section{Experiments}

\begin{table}[t]
    \centering
    \begin{adjustbox}{width=1.0\columnwidth,center}
    \begin{tabular}{l|c c c|c c c |c c c |c c c}
    \toprule
    & \multicolumn{3}{c|}{\textbf{19-1(2 stages)}} & \multicolumn{3}{c|}{\textbf{15-5(2 stages)}} & \multicolumn{3}{c}{\textbf{15-1(6 stages)}} \\
    Method & 0-19 & 20 & all & 0-15 & 16-20 & all & 0-15 & 16-20 & all \\
    \midrule
    Joint & 79.45 & 72.94 & 79.14 & 79.77 & 72.35 & 77.43 & 78.88 & 72.63 & 77.39\\
    \midrule
    EWC \cite{kirkpatrick2017overcoming} & 26.90 & 14.00 & 26.30 & 24.30 & 35.50 & 27.10 & 0.30 & 4.30 & 1.30\\
    LwF-MC \cite{rebuffi2017icarl} & 64.40 & 13.30 & 61.90 & 58.10 & 35.00 & 52.30 & 6.40 & 8.40 & 6.90 \\ 
    ILT \cite{michieli2019incremental}& 67.75 & 10.88 & 65.05 & 67.08 & 39.23 & 60.45 & 8.75 & 7.99 & 8.56 \\
    MiB \cite{cermelli2020modeling}& 70.57 & 22.82 & 68.30 & 75.30 & 48.68 & 68.96 & 39.47 & 14.50 & 33.53\\
    RCN \cite{zhang2022representation}& & & & 78.80 & 52.00 & 72.40 & 70.60 & 23.70 & 59.40 \\
    REMINDER \cite{phan2022class}& 76.48 & 32.34 & 74.38 & 76.11 & 50.74 & 70.07 & 68.30 & 27.23 & 58.52 \\
    SDR \cite{michieli2021continual}& 68.52 & 23.29 & 66.37 & 75.21 & 46.72 & 68.64 & 43.08 & 19.31 & 37.42\\
    PLOP \cite{douillard2021plop}& 75.35 & 37.35 & 73.54 & 75.73 & 51.71 & 70.09 & 65.12 & 21.11 & 54.64  \\
    \midrule
    Ours & 79.40 & 42.80 & 77.66 & 79.31 & 55.88 & 73.73 & 78.54 & 50.82 & 71.94\\
    \bottomrule
    \end{tabular}
    \end{adjustbox}
    \caption{Comparison results on Pascal-VOC 2012.}
    \label{pascal_compare}
\vspace{-1.5\baselineskip}
\end{table}

\begin{table}[]
    \centering
    \begin{adjustbox}{width=1.0\columnwidth,center}
    \begin{tabular}{l|c c c|c c c|c c c|c c c|c c c}
    \toprule
    & \multicolumn{3}{c|}{\textbf{100-50(2 stages)}} & \multicolumn{3}{c|}{\textbf{100-10(6 stages)}} & \multicolumn{3}{c|}{\textbf{100-5(11 stages)}}\\
    Method & 0-100 & 101-150 & all & 0-100 & 101-150 & all & 0-100 & 101-150 & all \\
    \midrule
    Joint & 44.34 & 28.21 & 39.00 & 44.34 & 28.21 & 39.00 & 44.34 & 28.21 & 39.00 \\
    \midrule
    ILT\cite{michieli2019incremental} & 18.29 & 14.40 & 17.00 & 0.11 & 3.06 & 1.09 & 0.08 & 1.31 & 0.49\\
    MiB\cite{cermelli2020modeling} & 40.52 & 17.17 & 32.79 & 38.21 & 11.12 & 29.24 & 36.01 & 5.66 & 25.96\\
    SDR \cite{michieli2021continual} & 37.40 & 24.80 & 33.20 & 12.13 & 28.94 & 34.48 & 33.02 & 10.63 & 25.61\\
    PLOP\cite{douillard2021plop} & 41.87 & 14.89 & 32.94 & 40.48 & 13.61 & 31.59 & 35.72 & 12.18 & 27.93 \\
    REMINDER\cite{phan2022class} & 41.55 & 19.16 & 34.14 & 38.96 & 21.28 & 33.11 & 36.06 & 16.38 & 29.54\\
    \midrule 
    Ours & 44.06 & 24.96 & 37.74 & 43.88 & 25.14 & 37.67 & 43.35 & 18.53 & 35.13\\
    \bottomrule

    \end{tabular}
    \end{adjustbox}
    \caption{Comparison results on ADE 20K.}
    \label{ade_compare}
\end{table}

\subsection{Comparisons with the State-of-the-arts}
 We compare the segmentation performance of our method with other state-of-the-art CSS methods on two datasets, including  Pascal-VOC 2012 and ADE 20K. 
The performance is evaluated with three metrics. The first one is the mIoU over the initial classes $\mathcal{C}_{1}$, and the second one measures the mIoU for all incremental classes $\mathcal{C}_{2:T}$. The third metric (all) denotes the mIoU for all observed classes $\mathcal{C}_{1:T}$. 
In experiments, We follow previous works \cite{douillard2021plop, phan2022class} by using Deeplab-v3 with the ResNet-101 backbone as the segmentation model. Following \cite{cha2021ssul}, the memory length $|\mathcal{M}|$ is 100 and 300 for Pascal-VOC 2012 and ADE20K, respectively. We adopt the widely-used pseudo label mechanism for training the segmentation network. Due to the paper length limitation, please see the \textbf{supplementary material} for more implementation details, segmentation model training details and visualization results. 

Table. \ref{pascal_compare} presents the performance on Pascal-VOC 2012 for three different settings including 19-1 (2 stages), 15-5 (2 stages) and 15-1 (6 stages). Our method achieves state-of-the-art performance. On the three settings, our method achieves 77.66\%, 73.73\%, and 71.94\% mIoUs on the `all' metric, outperforming the second-place method by 3.29\%, 1.33\%, and 12.54\%, respectively. The improvement is especially significant for the 15-1 (6 stages) setting, which is quite challenging due to the more severe catastrophic forgetting issue caused by a larger number of continuous stages. Our method, with carefully selecting and enhancing the replay samples, shows elevated effectiveness under such a challenging scenario. 

The comparison results with the ADE 20K are shown in Table. \ref{ade_compare}. For 3 different settings including 100-50 (2 stages), 100-10 (6 stages) and 100-5 (11 stages), our method achieves 37.74\%, 37.67\% and 35.13\% mIoUs on the `all' metric, improving the second-place one by 2.60\%, 4.56\% and 5.59\% respectively, showing its effectiveness and advantage.

\begin{table}[t]
    \centering
    \footnotesize
    \begin{tabular}{l| c c c}
    \toprule
    Selection Strategy & 0-15 & 16-20 & all\\
    \midrule
     Random Selection & 72.82 & 32.21 & 63.15\\
     \midrule
     iCaRL \cite{rebuffi2017icarl} & 73.91 & 39.11 &  65.62\\
     Rainbow \cite{bang2021rainbow} & 74.03 & 40.70 & 66.09\\
     CBES \cite{yan2021framework} & 74.15 & 41.57 & 66.39\\
     SSUL \cite{cha2021ssul} & 74.20 & 41.33 & 66.37\\
     \midrule
     NHS & 74.50 & 42.25 & 66.82\\
     \midrule 
     Ours (w/o Enhancement) & 77.54 & 45.98 & 70.02\\
     \bottomrule
    \end{tabular}
    \caption{Comparison with other sample selection strategies. NHS denotes  a new-designed hand-crafted strategy using the same factors as our method (sample diversity and class performance). For a fair comparison, we report the result of our method w/o the enhancement operation. }
    \label{selection_ablation}
\end{table}
\subsection{Comparison with Other Sample Selection Strategies}

\vspace{-0.2\baselineskip}
To verify the effectiveness of our RL-driven automatic replay mechanism, we validate and compare it with other sample selection methods in the CSS task. The experiments are conducted on Pascal-VOC 2012 under the 15-1 (6 stages) setting. The results are shown in Table.\ref{selection_ablation}. The compared methods include three types: 1) the random selection strategy; 2) the previously-proposed hand-crafted strategies including iCaRL \cite{rebuffi2017icarl}, Rainbow \cite{bang2021rainbow}, CBES \cite{yan2021framework} and SSUL \cite{cha2021ssul}.  Both iCaRL and Rainbow are diversity-based selection criteria. CBES and SSUL are two class-balanced sample selection strategies that are specially designed for CSS. Besides, to validate the effectiveness of the automatic learning mechanism, we also design a new hand-crafted strategy using the same factors as our method (sample diversity and class performance). The newly-designed one is based on our visualization of the learned policy introduced in Sec. \ref{analysis}. It shows selecting the common samples is effective for the hard classes with bad performance, while selecting the diverse samples is better for the simple classes with good performance. Thus, we design a strategy where the most common samples with the lowest diversity scores are selected for the top 50\% low-performance classes, while the most diverse samples with the highest diversity scores are selected for other high-performance classes. We denote the new-designed (N) hand-crafted (H) strategy (S) as NHS. On the `all' metric, random selection achieves 63.15\% mIoU. By smartly selecting the appropriate samples based on heuristic rules, iCaRL, Rainbow CBES and SSUL achieve 65.62\%, 66.09\% and 66.39\% and $66.37\%$ mIoUs, respectively, and NHS further improves it to 66.82\% by considering more factors with the complicated relationship. Considering these methods only select samples, for a fair comparison, we report the result of our method w/o the enhancement operation. It achieves 70.02\% mIoU,  not only outperforms the previously-proposed iCaRL, Rainbow, CBES and SSUL, showing the elevated effectiveness of the novel selection approach; but also outperforms NHS using the same set of factors, demonstrating the significant advantages of the reward-driven automatic policy learning mechanism over the hand-crafted strategies.

\begin{table}[t]
    \centering
    \begin{adjustbox}{width=0.85\columnwidth,center}
    \begin{tabular}{l | c c c}
    \toprule
    Method & 0-15 & 16-20 & all\\
    \midrule
    Ours & 78.54 & 50.82 & 71.94\\
    \midrule
    Ours w/o Enhancement & 77.54 & 45.98 & 70.02\\
    Ours w/o Enhancement \& Selection & 72.82 & 32.21 & 63.15\\ 
     \bottomrule
    \end{tabular}
    \end{adjustbox}
    \caption{Ablation results of the selection-enhancement dual-stage action.}
    \label{component_ablation}
\end{table}

\begin{table}[t]
    \centering
    \begin{adjustbox}{width=0.85\columnwidth,center}
    \begin{tabular}{l| c c c}
    \toprule
    Method & 0-15 & 16-20 & all\\
    \midrule
    Ours & 78.54 & 50.82 & 71.94\\
    \midrule
    Ours w/o $div$ & 74.09 & 33.33 & 64.39 \\
    Ours w/o $I$ & 76.50 & 42.08 & 68.30 \\
    Ours w/o $g$ & 77.79 & 45.32 & 70.06\\
    Ours w/o $\{I, g\}$ & 76.18 & 36.03 & 66.68\\
    \midrule
    Ours w/o $div$ w/ $div\_prototype$ & 76.93 & 47.16 & 69.83 \\
     \bottomrule
    \end{tabular}
    \end{adjustbox}
    \caption{Ablation results of the state representations.}
    \vspace{-0.7\baselineskip}
    \label{state_ablation}
\end{table}
\vspace{-0.5\baselineskip}
\subsection{Ablation Study}
\vspace{-0.5\baselineskip}
In this part, we perform ablation study to verify the effectiveness of different components in our method. All experiments are conducted on Pascal-VOC 2012 under the 15-1 (6 stages) setting. Due to the paper length limitation, more results including the ablation for memory length $|\mathcal{M}|$ and superpixel number $M$ are presented in supplementary materials.

\noindent \textbf{Ablation of Selection-enhancement Dual Stage Action.}
We conduct experiments to verify the effectiveness of the proposed selection-enhancement dual-stage action paradigm, with results shown in Table. \ref{component_ablation}. Our method with both the sample selection and enhancement actions achieves 71.94\% mIoU on the `all' metric. By removing the enhancement operation, the performance decreases to 70.02\%. By further removing both enhancement and selection procedures so that the memory is randomly filled, the performance is only 63.15\%, 8.79\% lower than our method. The results indicate that both the selection and enhancement operations can effectively boost CSS performance.


\noindent \textbf{Ablation of State Representation Design. }We then validate different components of the designed state representations and the results are presented in Table. \ref{state_ablation}. The state representation contains three parts: 1) sample diversity $div$; 2) accuracy $I$ and 3) forgetfulness $g$. The latter two constitute the class performance feature. In addition to validating the three parts, we also test using a common diversity metric instead of our novel one. Such a metric measures the inter-sample similarity by directly computing the distance between their prototype features. We name it as $div\_prototype$. Using $div$ shows significant performance improvement (69.83\% $\rightarrow$ 71.94\%) to $div\_common$, demonstrating the effectiveness of our novel graph-based similarity. 

\vspace{-0.5\baselineskip}
\subsection{Analysis of the Learned Policy}\label{analysis}
\begin{figure}[t]
    \centering
    \includegraphics[width=0.7\linewidth]{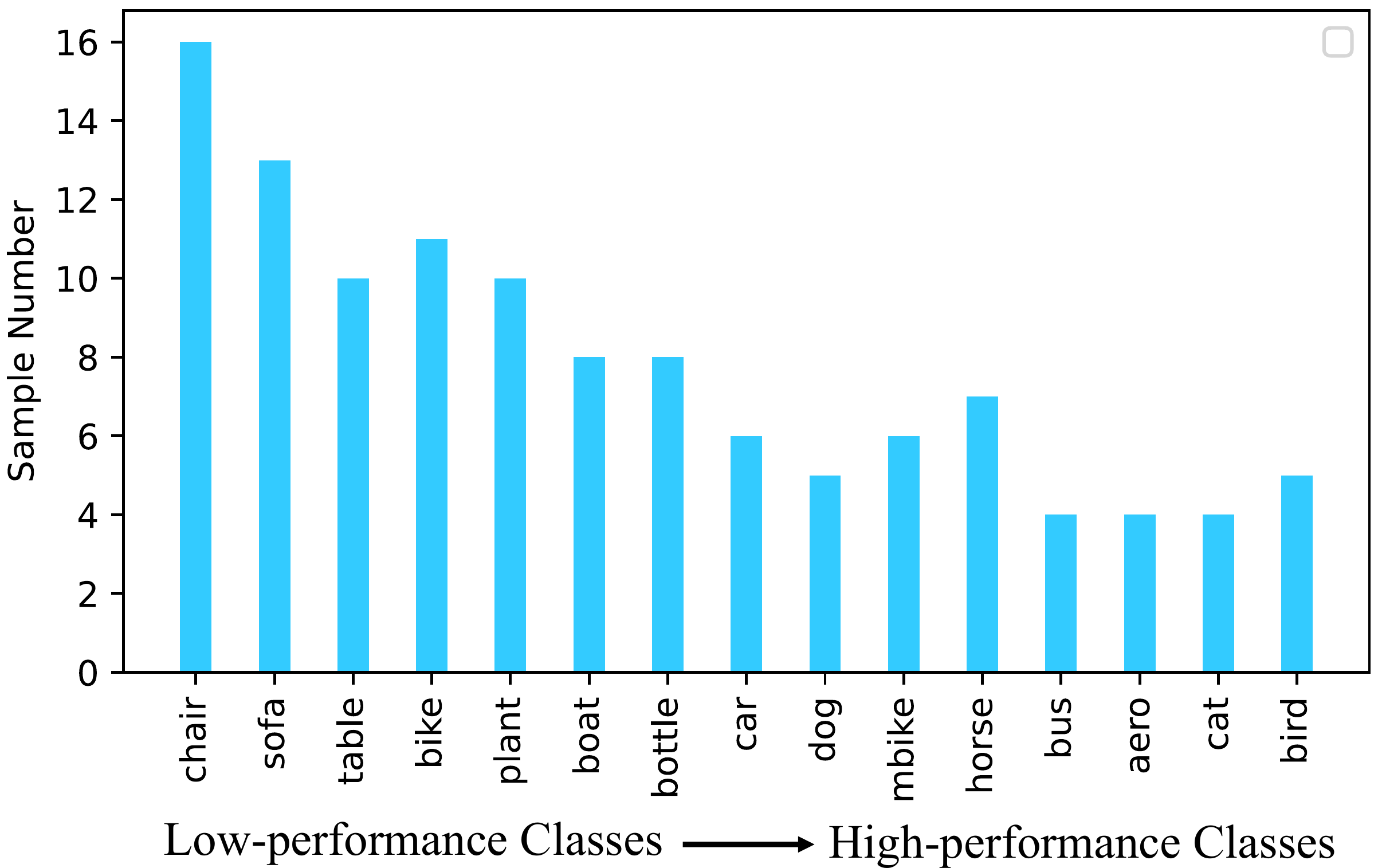}
    \caption{The numbers of selected samples for different classes. The horizontal axis from left to right represents classes from poor to good performance.}
    \label{ablation_number}
\vspace{-0.8\baselineskip}
\end{figure}

\vspace{-0.5\baselineskip}
We further analyze the learned sample selection policy both qualitatively and quantitatively to offer more insights into how our method works. After analyzing the learned policy, we can observe the following rules:

\textbf{(1) Low-performance classes require more replay samples.} As shown in Fig.\ref{ablation_number}, on Pascal-VOC 2012 dataset, we count the number of selected samples for different classes with different performances. From left to right, the horizontal axis represents the classes from low to high performance. We can find the negative correlation between class performance and the selected sample number. The low-performance classes are less accurate or more easily to be forgotten, so more samples are required for replay to alleviate the more severe catastrophic
forgetting issue. 

\begin{figure}[t]
	\centering
	\begin{subfigure}{0.33\linewidth}
		\centering
		\includegraphics[width=0.9\linewidth]{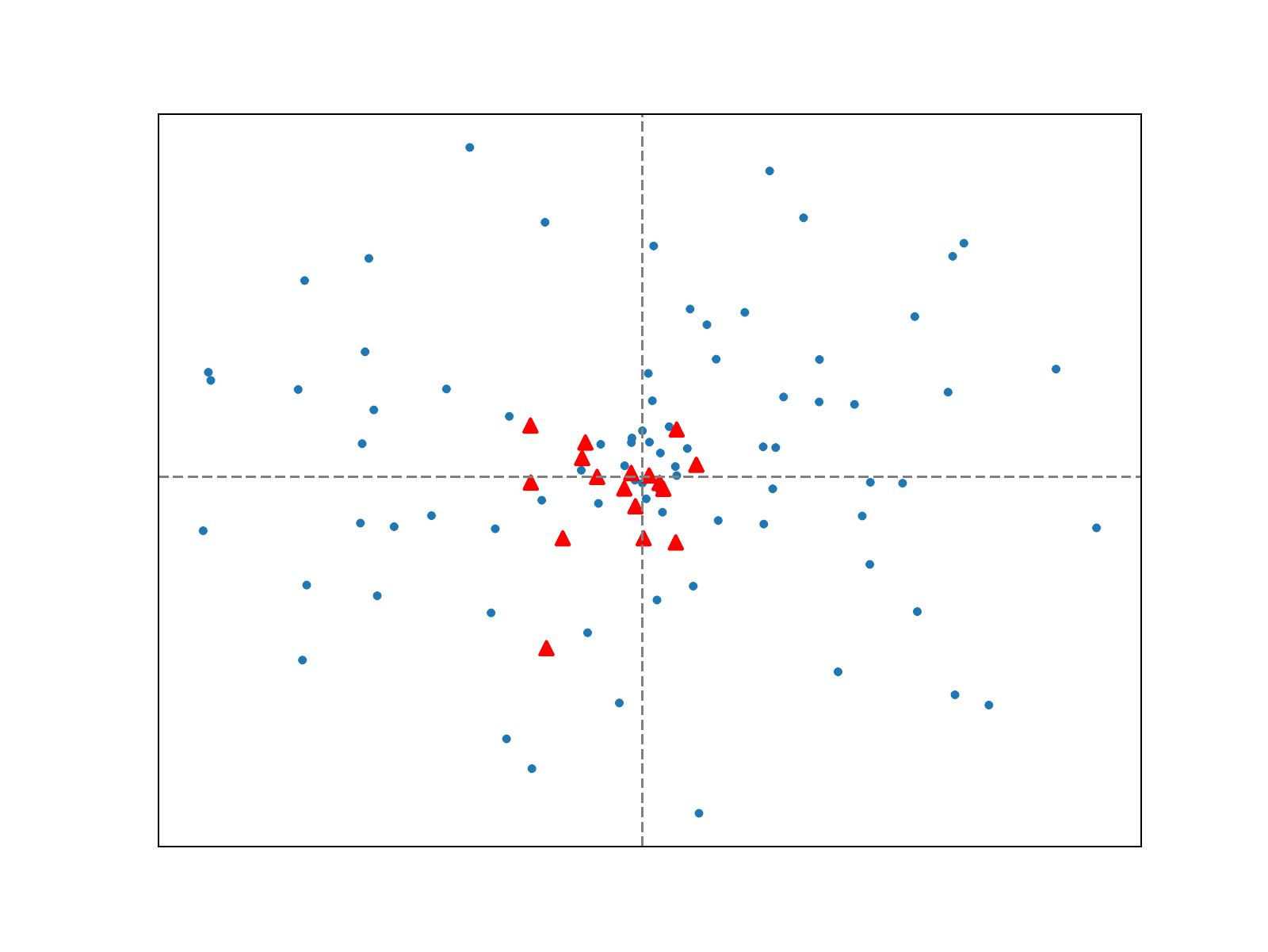}
		\caption{Chair}
		\label{chair_div}
	\end{subfigure}
    \centering
    \begin{subfigure}{0.33\linewidth}
		\centering
    	\includegraphics[width=0.9\linewidth]{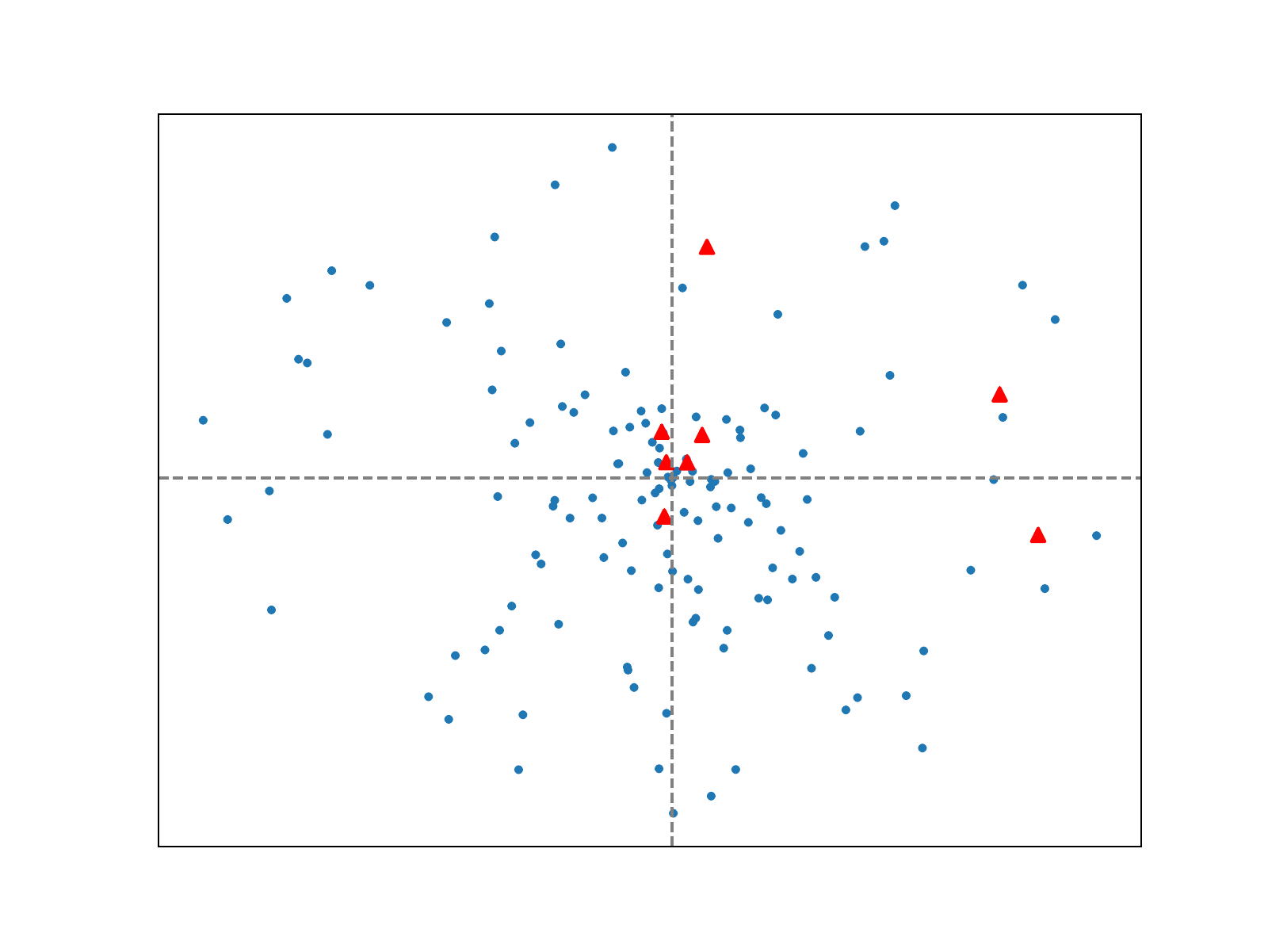}
		\caption{Boat}
		\label{boat_div}
	\end{subfigure}
	    \centering
    \begin{subfigure}{0.33\linewidth}
		\centering
    	\includegraphics[width=0.9\linewidth]{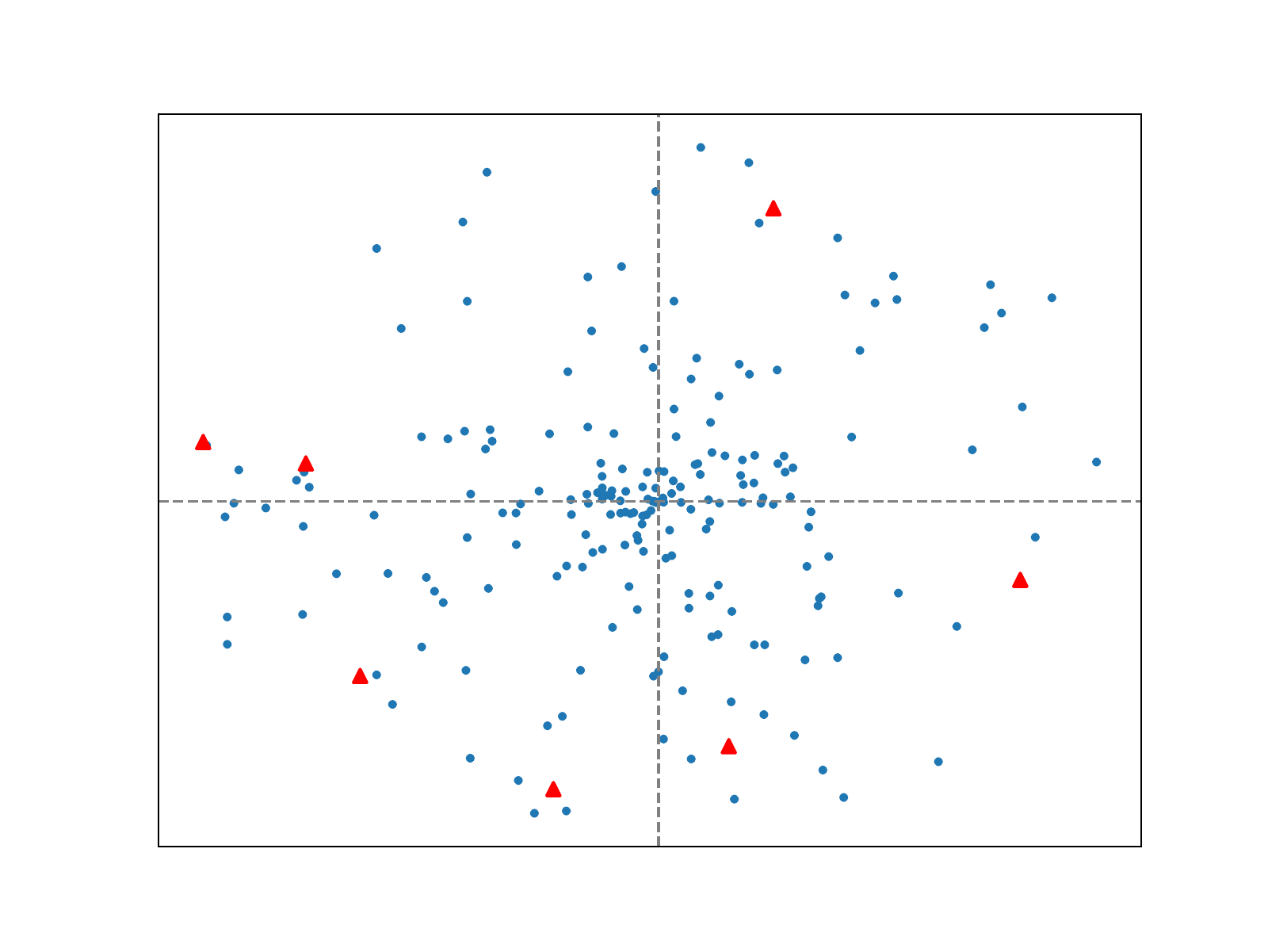}
		\caption{Bird}
		\label{bird_div}
	\end{subfigure}
	\caption{(Best viewed in color). Visualization of the diversity for the selected samples of three classes including `chair', `boat' and `bird'. The {\color{red} red} triangles represent the selected samples and the {\color{blue} blue} dots denote other samples that are not selected. Triangles or dots closer to the center represent samples with the lower diversity.}

	\label{class_div}
\end{figure}

\textbf{(2) Classes with different performances require different kinds of samples.} 
We further investigate the learned strategy for classes with different performances. 
We visualize the diversity of the selected samples for three representative classes: `chair', `boat', and `bird.'  `chair' is a hard class with a low class performance, `bird' is an easy class with a high class performance, and `boat' has a medium class performance. The results are shown in Fig. \ref{class_div}, where the red triangles represent the selected samples, and the blue dots denote other samples that are not selected. Triangles or dots closer to the center represent samples with lower diversity. As can be observed, for the low-performance class `chair', most red triangles are distributed in the center, indicating the agent selects common samples with low diversity. 
On the contrary, for the high-performance class `bird', the high-diversity samples are selected.
For the middle-performance class `boat', both the common and diverse samples are selected. We believe the different degrees of forgetfulness for different classes can explain the learned policy. For hard classes where the catastrophic forgetting is more severe, most samples including both the high-diversity novel ones and low-diversity common ones are  forgotten after the model trains on new classes, so using the more common and representative samples can learn a classification space covering most samples. On the contrary, for easy classes with relatively minor catastrophic forgetting issues, the common samples can still be remembered in the next stage while the high-diversity samples are easier to be forgotten. Thus, replay with high-diversity samples can be more effective.  

\vspace{-0.5\baselineskip}
\section{Complexity Discussion}
\vspace{-0.5\baselineskip}
\label{complexity}
Training the agent network requires additional time. According to Alg. \ref{alg}, the theoretically additional cost is $O\left(Y\right)$ higher than the time for deployment. However, we argue that agent training is an offline process and we can use a shallower segmentation network and a smaller dataset for training. With these simplifications, we get a computation-efficient agent training process where the training time is 1.16 times that of the deployment phase for the 15-5 (2 stages) setting on Pascal-VOC 2012. Also, the agent trained on one dataset can be deployed on other datasets. Thus the agent only needs to be trained once and can be deployed to different CSS tasks. We present the experimental details for such a cross-dataset deployment in supplementary materials. The additional cost for using the agent in the deployment phase is minor (8.23\% and 12.96\% of the total training time on Pascal-VOC 2012 and ADE 20K, respectively). 
\vspace{-0.5\baselineskip}
\section{Conclusion}
\vspace{-0.5\baselineskip}
In this work, we propose a novel and automatic memory selection paradigm. It significantly facilitates alleviating the severe catastrophic forgetting issue through more effective memory management in the Continual Semantic Segmentation (CSS) task. We propose a novel learning-based approach with an agent network to automatically learn the policy. The input representation to the agent network is tailored for the CSS task. We also use the agent network to further perform a novel sample enhancement operation through a gradient-based approach to boost the effectiveness of selected samples. The work provides valuable insights into the memory selection of continual semantic segmentation and practical tools that is readily applicable. Our method is effective and general, as shown by our extensive experiments with state-of-the-art (SOTA) performance.
\\

\noindent \textbf{Acknowledgement}
This research is supported by the National Research Foundation, Singapore under its AI Singapore Programme (AISG Award No: AISG2-PhD-2021-08-006), MOE AcRF Tier 2 (Proposal ID: T2EP20222-0035) and SUTD SKI Project (SKI 2021\_02\_06).

{\small
\bibliographystyle{ieee_fullname}
\bibliography{egbib}
}
\clearpage

\appendix
\section{Details for Inter-graph Similarity Computation}
Here we present the details for computing the inter-graph similarity through Sikhorn algorithm. Considering a graph $\mathcal{G} = \{F_{m}, \{D^{m, n}\}_{n=1}^{M}\}_{m=1}^{M}$, we first generate a single-vector representation for each vertex by aggregating other vertices. The aggregation is achieved through weighted sum written as:
\begin{equation}
    \hat{F}_{m} = \frac{1}{\sum_{n=1}^{M}W_{m, n}}\sum_{n=1}^{M}W_{m, n}F_{n},
\end{equation}
Where the weight $W_{m, n}$ is formulated by:
\begin{equation}
    W_{m, n} = \exp\left(-{D}^{m, n}\right).
\end{equation}
In this way, $\mathcal{G}$ is represented as $\{\hat{F}_{m}\}_{m=1}^{M}$. For simplify, we denote $\mathcal{G}_{i}$ for the $i$-th image as $\{\hat{F}^{i}_{m}\}_{m=1}^{M}$. Next, we match $\mathcal{G}_{i}$ and $\mathcal{G}_{j}$ by solving an optimal transport (OT) task:
\begin{equation}
    \mathop{\rm Min}\limits_{A}\sum\limits_{a,b}A_{a, b}M_{a,b},
\end{equation}
where $A$ is the transportation plan that implies the alignment information and $M$ is the cost matrix. $M_{a, b}$ measures the transport cost from the $a$-th vertex $\hat{F}^{i}_{a}$ in $\mathcal{G}_{i}$ to the $b$-th vertex $\hat{F}^{j}_{b}$ in $\mathcal{G}_{i}$, which is written as:
\begin{equation}
    M_{a, b} = 1 - {\rm Cos}\left(\hat{F}^{i}_{a}, \hat{F}^{j}_{b}\right),
\end{equation}
where ${\rm Cos}$ denotes the cosine similarity. The unique solution $A^{*}$ can be calculated through Sinkhorn’s algorithm:
\begin{equation}
    A^{*} = {\rm diag}\left(\mathbf{u}\right)K{\rm diag}\left(\mathbf{v}\right),
\end{equation}
where the vectors $\mathbf{u}$ and $\mathbf{v}$ are obtained through the above iterations:
\begin{equation}
\begin{aligned}
    &\mathbf{v}^{t=0} = \frac{\mathbf{1}_{m}}{\mathbf{v}^{t+1}},\\
    &\mathbf{u}^{t+1},~\mathbf{v}^{t+1} = \frac{\mathbf{1}_{n}}{K\mathbf{u}^{t+1}}
\end{aligned}
\end{equation}
we set the iteration number to be 5. Finally, the transport cost $tc$ is computed as:
\begin{equation}
    tc = \sum\limits_{a,b}A^{*}_{a, b}M_{a,b},
\end{equation}
which measures the similarity between $\mathcal{G}_{i}$ and $\mathcal{G}_{j}$

\section{Implementation Details} 
Our method contains two phases: agent training and policy deployment. The first phase trains the agent network to get the selection policy, while the latter phase employs the trained policy for the CSS training. 

For the deployment phase, the hyper-parameters settings follow the previous work \cite{douillard2021plop}. Concretely, we adopt SGD as the optimizer, where the momentum value is 0.9 and the initial learning rate is 1e-2 with the `poly' learning rate decay schedule. For each continual stage, the network is trained for 30 epochs on Pascal VOC and 60 epochs for ADE20K. The batch size is 24 for both datasets. Following \cite{cha2021ssul}, the memory length $|\mathcal{M}|$ is 100 and 300 for Pascal-VOC 2012 and ADE20K, respectively. Following \cite{li2021adaptive}, the superpixel number $M$ for computing sample diversity is 5, $\epsilon$ in Eq. 5 is 0.1. 

For the agent training phase, as we have discussed in Sec. 6 of text, we use the different hyper-parameters settings to speed up training. Concretely, in this phase, we use Deeplabv3 with ResNet18 backbone as the segmentation model. The training epochs $Y$ in Alg. 1 is 1000. We randomly partition 10\% of whole data into the training set and leave others as the reward set. For each continual stage, the network is trained for 5 epochs on Pascal VOC and 8 epochs for ADE20K. The segmentation network is optimized by SGD with the initial rate being 0.01, and the agent network is optimized by Monmentumn with the learning rate being 0.1. 

\section{Segmentation Training}
In each stage $t$ of a CSS task, both the memory $\mathcal{M}$ and current dataset $\mathcal{D}_{t}$ are utilized for training the segmentation model. We follow previous works by using the widely-adopted pseudo-label mechanism to enhance the segmentation training performance. Concretely, the pixels belonging to previous and future classes become the background for images in the current stage. Considering the model is trained on the combined ground truth from both current and previous classes, we use its prediction to generate pseudo labels for background pixels for images in $\mathcal{M}$ and $\mathcal{D}_{t}$. Concretely, let's denote 0 be the background class. For a sample $X$ with the ground truth label $Y$, we first use the current segmentation model to get its prediction mask $P$ and the confidence map $M$, then the pseudo ground truth label $\hat{Y}^{t}_{i}$ for the $i$-th pixel on $X$ is obtained by:
\begin{equation}
    \hat{Y}_{i}^{t} = \left\{
    \begin{aligned}
    Y_{i},\quad &{\rm if}\ Y_{i} \neq 0\\
    P_{i},\quad &{\rm if}\ Y_{i} = 0\ {\rm and}\ M_{i} > 0.8\\
    0,\quad &{\rm else}
    \end{aligned}
    \right.
\end{equation}
Eventually, $X$ along with the generated pseudo label $\hat{Y}^{t}$ are used for training the segmentation model through the cross-entropy loss.  

\section{More Ablation Results}
\begin{figure}[t]
    \centering
    \includegraphics[width=0.7\linewidth]{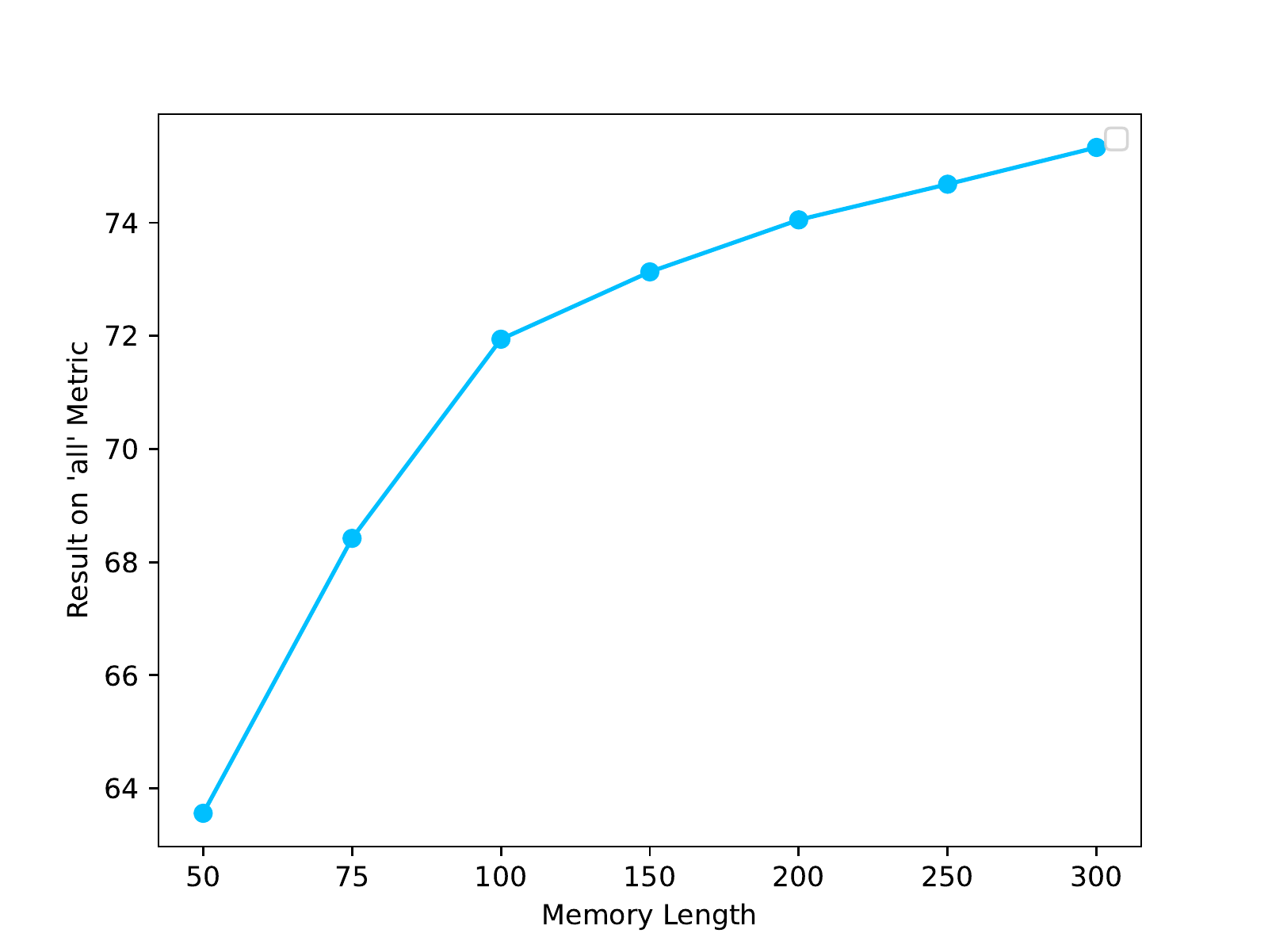}
    \caption{Ablation results of memory length. As the memory length increases from 50 to 300, the mIoU on the `all' metric increases from 59.10 to 68.37.}
    \label{ablation_length}
\end{figure}

\noindent \textbf{Ablation of Memory Length.} In the experiment section, for the fair comparison, we follow \cite{cha2021ssul} by setting the memory length $\mathcal{M}$ to 100 and 300 for Pascal-VOC 2012 and ADE20K, respectively. We further validate the performance for the 15-1(6 stages) setting on Pascal-VOC 2012 by using memories with different lengths ranging from 50 to 300. The results are shown in Fig. \ref{ablation_length}. We can observe that a larger memory brings better performance. As the memory length increases from 50 to 300, the mIoU on the `all' metric increases from 63.56 to 75.33. \\

\begin{figure}[t]
    \centering
    \includegraphics[width=0.7\linewidth]{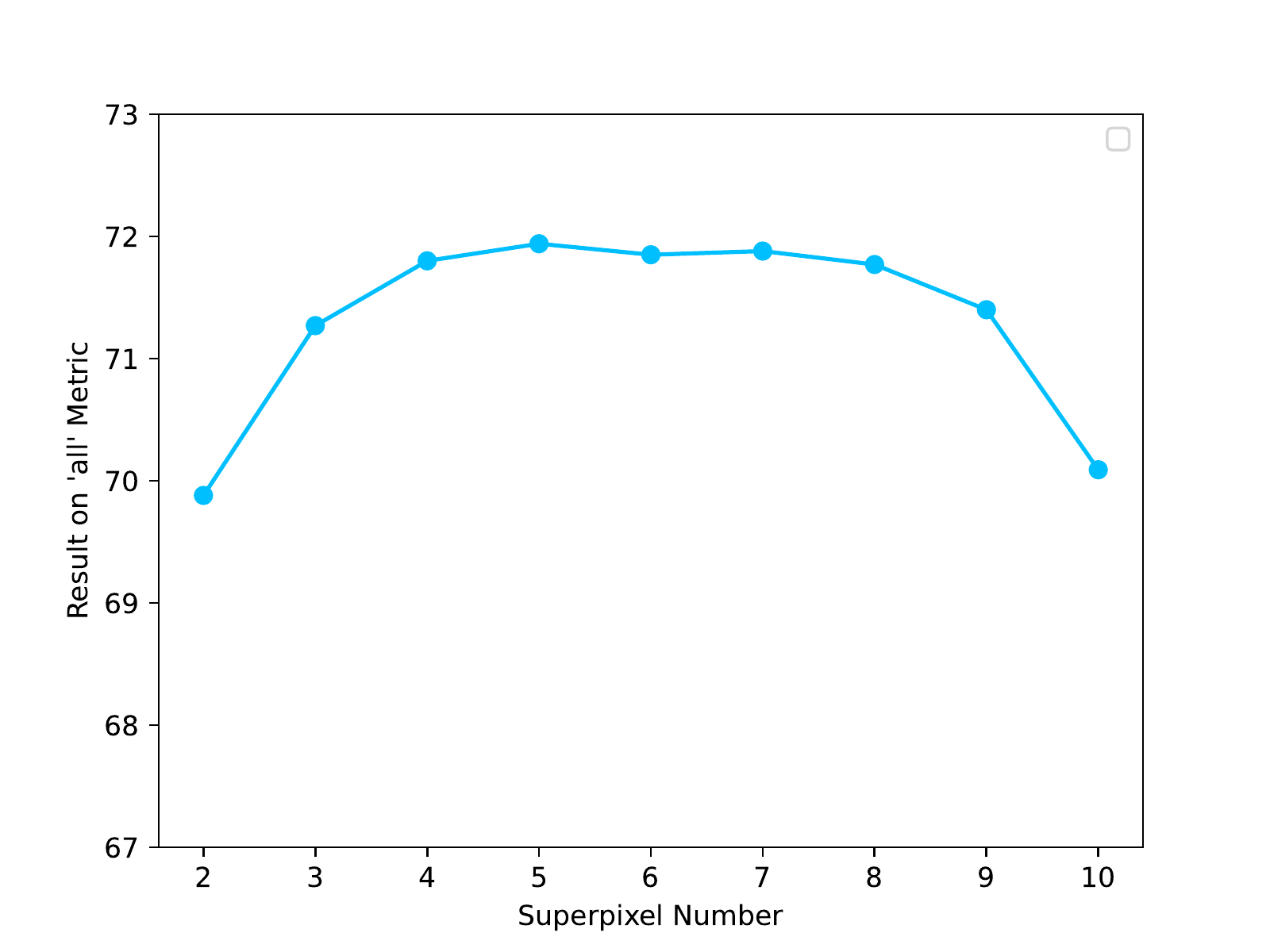}
    \caption{Ablation results of superpixel number. }
    \label{ablation_superpixel}
\end{figure}

\noindent \textbf{Ablation of Superpixel Number.}
In order to compute the sample diversity, each region is divided into $M$ superpixels for constructing the graph. Here we perform experimenters to validate how $M$ affects the performance and present the results in Fig. \ref{ablation_superpixel}. As can be observed, the performance keeps stable when $M$ is larger than 3 and smaller than 9, while a too large $M$ leads to the over segmenting that negatively affects the performance to some extent. Generally speaking, our method is non-sensitive to the hyper-parameter $M$, demonstrating its high robustness. 

\section{Discussion of State Representation Computation}
As illustrated in Line 435, Sec. 4.2.1 of the text, for computing the sample diversity $div$ and forgetfulness $g_{c}$, we introduce a support set $\mathcal{S}_{c}$ for each class $c$ that contains several graphs for images within $c$. To relieve the computation burden, for each current class in $\mathcal{C}_{t}$ that has a larger number of samples, we randomly sample 10\% from all images to form $\mathcal{S}_{c}$. Then sample diversity is derived by computing and averaging the inter-graph similarities with all graphs in $\mathcal{S}_{c}$. We conduct experiments on the 15-1 (6 stages) setting for Pascal-VOC 2012 dataset to verify the effectiveness. Loading all images into $\mathcal{S}_{c}$ gets 72.25\% mIoU for the `all' metric, which is just slightly better than the sampled set, which achieves 71.94\% mIoU. However, computing similarities on all images consumes 10 times more time than using the sampled set, which is unacceptable. Therefore, our strategy can be computationally efficient yet effective. 
\begin{figure}[t]
	\centering
    \begin{subfigure}{0.45\linewidth}
		\centering
    	\includegraphics[width=0.9\linewidth]{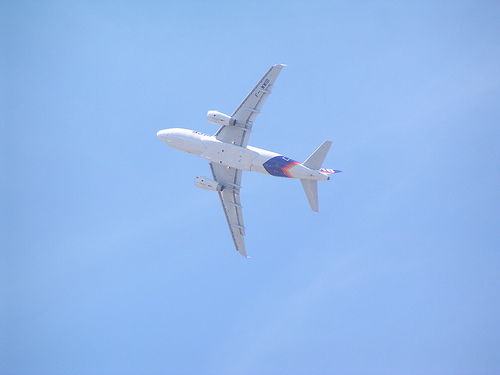}
	\end{subfigure}
	    \centering
    \begin{subfigure}{0.45\linewidth}
		\centering
    	\includegraphics[width=0.9\linewidth]{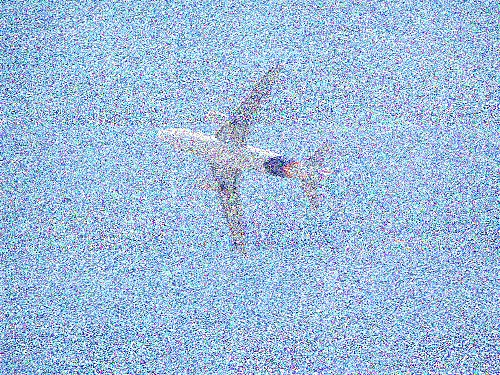}
	\end{subfigure}\\
	\begin{subfigure}{0.45\linewidth}
		\centering
    	\includegraphics[width=0.9\linewidth]{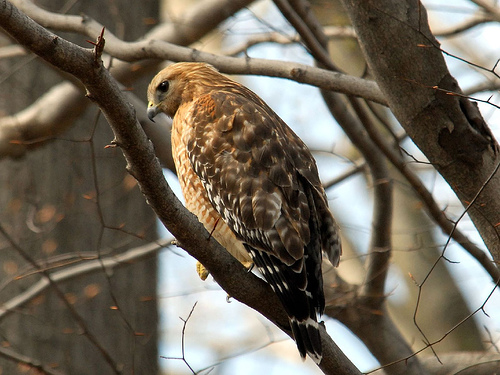}
	\end{subfigure}
	    \centering
    \begin{subfigure}{0.45\linewidth}
		\centering
    	\includegraphics[width=0.9\linewidth]{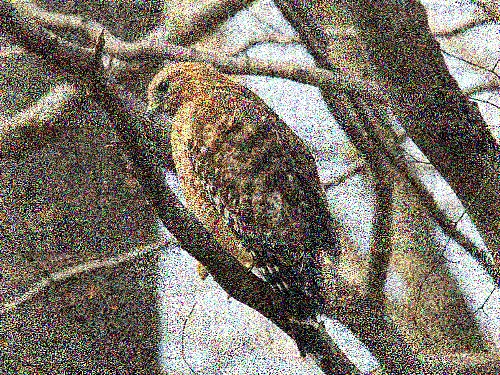}
	\end{subfigure}\\
	\begin{subfigure}{0.45\linewidth}
		\centering
    	\includegraphics[width=0.9\linewidth]{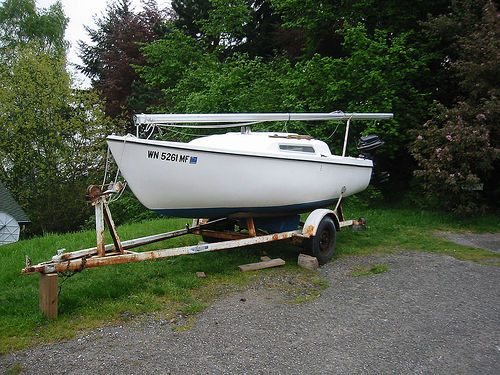}
    	\caption{original Image}
	\end{subfigure}
	    \centering
    \begin{subfigure}{0.45\linewidth}
		\centering
    	\includegraphics[width=0.9\linewidth]{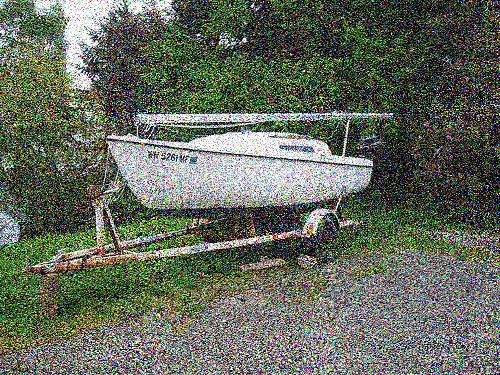}
    	\caption{Enhanced Image}
	\end{subfigure}
	
	\caption{Comparison between the original images and images after enhacement.}
	\label{enhance}
\end{figure}

\begin{figure}[t]
    \centering
    \includegraphics[width=0.8\linewidth]{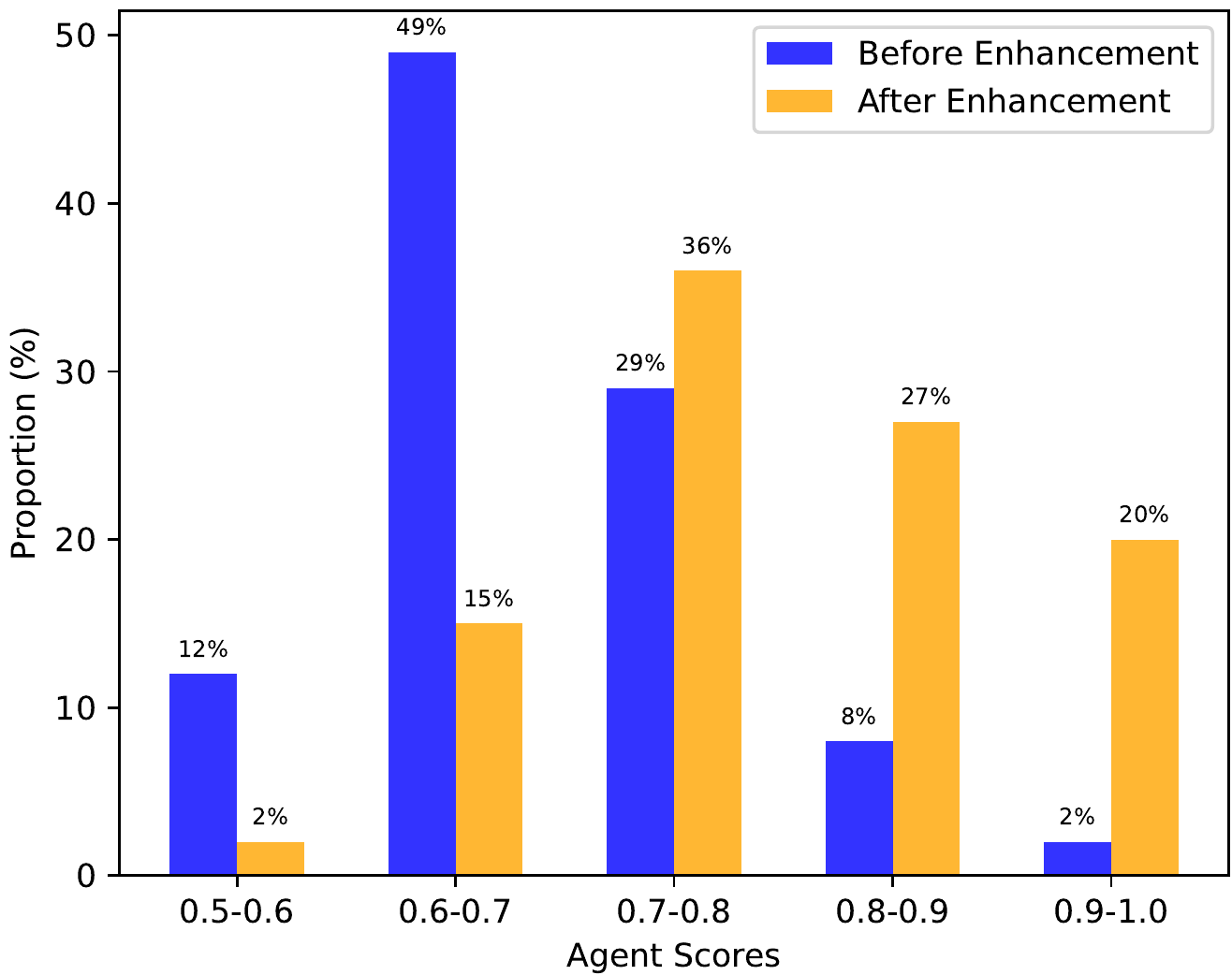}
    \caption{Comparison of agent score distributions for all selected samples before and after enhancement. The horizontal axis represents different score intervals. The vertical axis indicates the proportion of samples falling into each interval.}
    \label{proportion}
\end{figure}

\section{Cross-dataset Deployment} As discussed in Sec. 6 of the text, we can use an agent trained on one dataset to deploy on other datasets. We perform experiments to verify that capability. For the `all' metric, using the agent trained on Pascal-VOC 2012 to deploy on the 100-50(2 stages) setting of ADE 20K achieves 34.87\% mIoU, and using the agent trained on ADE 20K to deploy on 19-1(2 stages) setting of Pascal-VOC 2012 achieves 74.96\% mIoU, with both cases showing good performance. The results demonstrate the high generalization of our method. In realistic applications, the agent only needs to be trained once and then can be used on several different CSS tasks without the extra computation cost for agent retraining.

\section{Visualization of Sample Enhancement}
Our method includes a novel enhancement action. It enables the selected samples to have the better replay effectiveness by maximizing their agent scores through gradient-based editing. We presents some comparison results between original images and the enhanced images in Fig. \ref{enhance}. We also provide a quantitative comparison in Fig. \ref{proportion} to show the agent score distributions for all selected samples before and after enhancement, where the horizontal axis represents different score intervals, and the vertical axis indicates the proportion of samples falling into each interval. We can observe that after enhancement, there are more samples with high agent scores. This demonstrates that the gradient-based enhancement effectively increases agent scores, thus promoting the replay performance.

\section{Visualization of Segmentation Results}
\begin{figure*}[t]
	\centering
    \begin{subfigure}{0.24\linewidth}
		\centering
    	\includegraphics[width=0.9\linewidth]{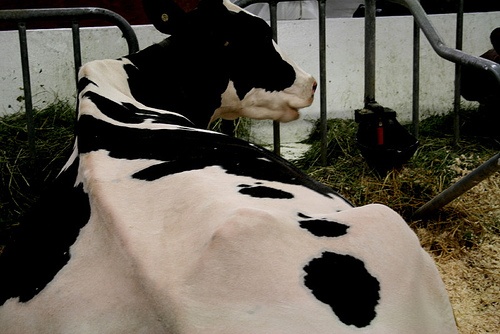}
	\end{subfigure}
	    \centering
    \begin{subfigure}{0.24\linewidth}
		\centering
    	\includegraphics[width=0.9\linewidth]{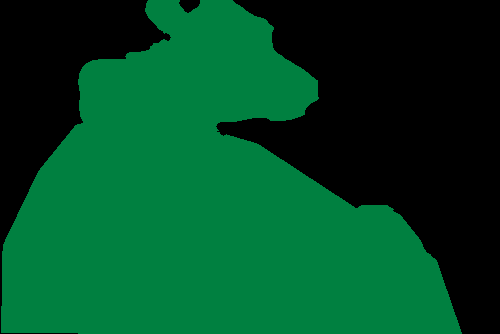}
	\end{subfigure}
		\centering
    \begin{subfigure}{0.24\linewidth}
		\centering
    	\includegraphics[width=0.9\linewidth]{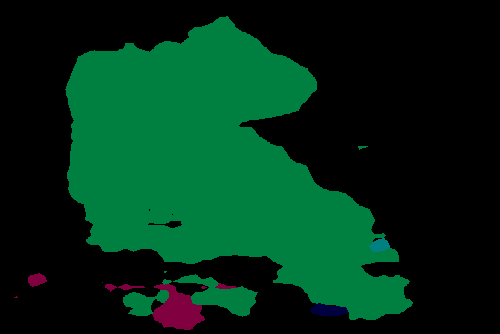}
	\end{subfigure}
	\begin{subfigure}{0.24\linewidth}
		\centering
    	\includegraphics[width=0.9\linewidth]{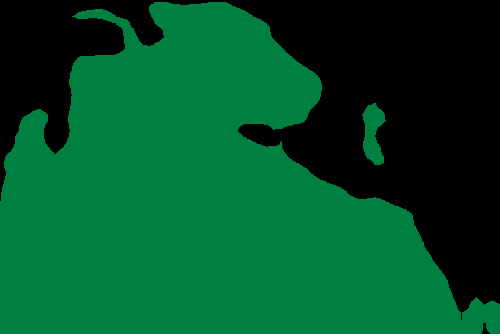}
	\end{subfigure}\\
	\centering
    \begin{subfigure}{0.24\linewidth}
		\centering
    	\includegraphics[width=0.9\linewidth]{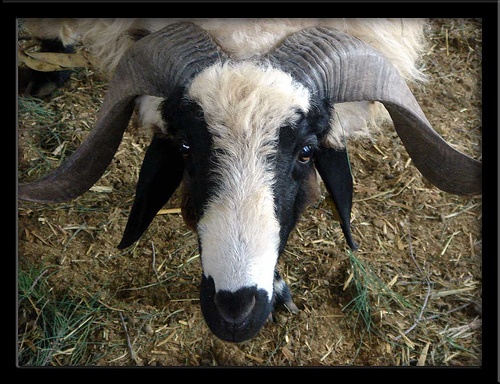}
	\end{subfigure}
	    \centering
    \begin{subfigure}{0.24\linewidth}
		\centering
    	\includegraphics[width=0.9\linewidth]{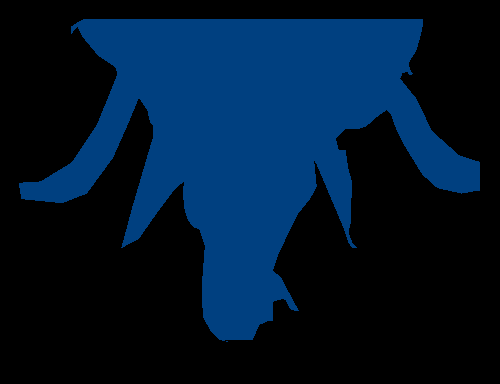}
	\end{subfigure}
		\centering
    \begin{subfigure}{0.24\linewidth}
		\centering
    	\includegraphics[width=0.9\linewidth]{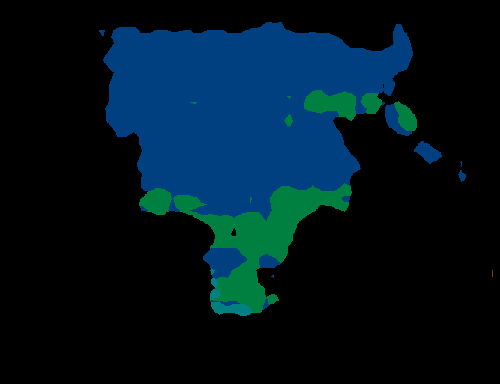}
	\end{subfigure}
	\begin{subfigure}{0.24\linewidth}
		\centering
    	\includegraphics[width=0.9\linewidth]{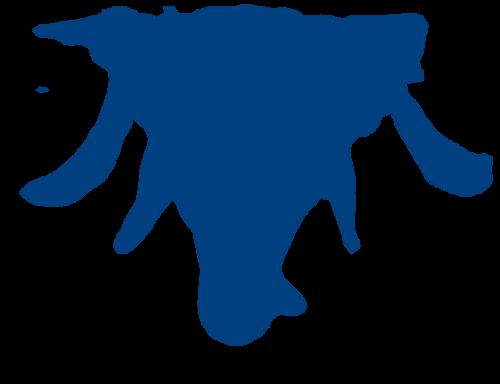}
	\end{subfigure}\\
	\centering
    \begin{subfigure}{0.24\linewidth}
		\centering
    	\includegraphics[width=0.9\linewidth]{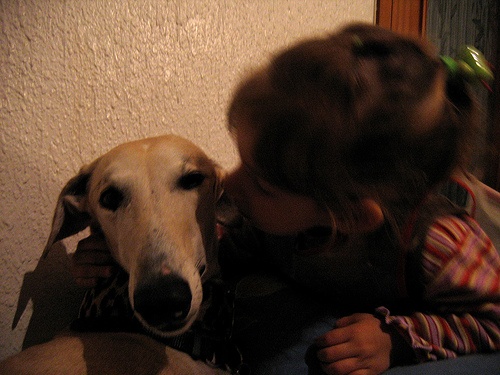}
	\end{subfigure}
	    \centering
    \begin{subfigure}{0.24\linewidth}
		\centering
    	\includegraphics[width=0.9\linewidth]{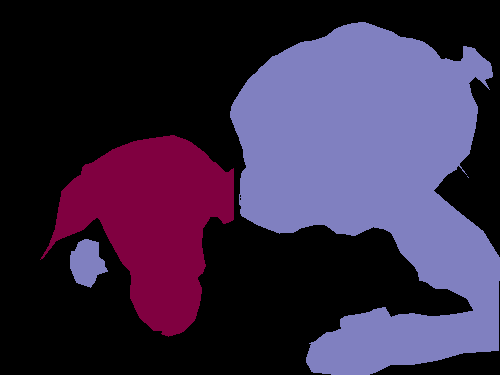}
	\end{subfigure}
		\centering
    \begin{subfigure}{0.24\linewidth}
		\centering
    	\includegraphics[width=0.9\linewidth]{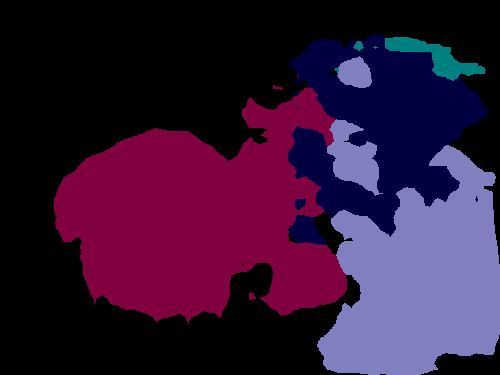}
	\end{subfigure}
	\begin{subfigure}{0.24\linewidth}
		\centering
    	\includegraphics[width=0.9\linewidth]{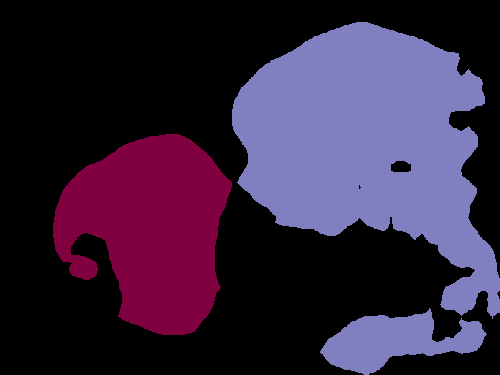}
	\end{subfigure}\\
	\centering
    \begin{subfigure}{0.24\linewidth}
		\centering
    	\includegraphics[width=0.9\linewidth]{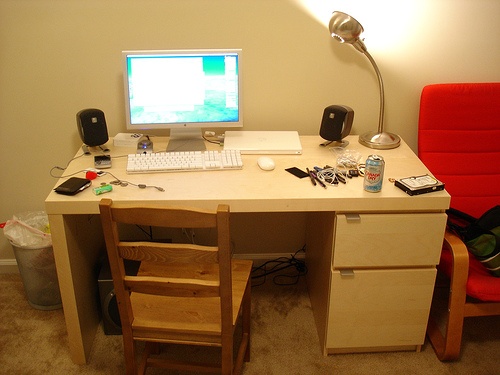}
	\end{subfigure}
	    \centering
    \begin{subfigure}{0.24\linewidth}
		\centering
    	\includegraphics[width=0.9\linewidth]{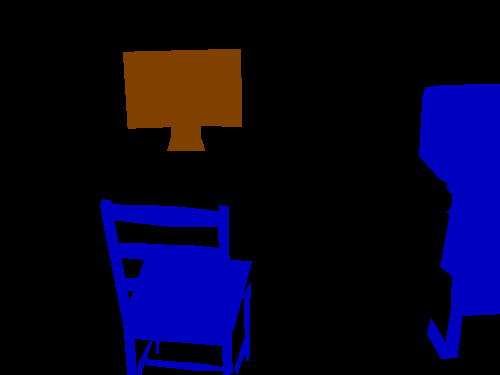}
	\end{subfigure}
		\centering
    \begin{subfigure}{0.24\linewidth}
		\centering
    	\includegraphics[width=0.9\linewidth]{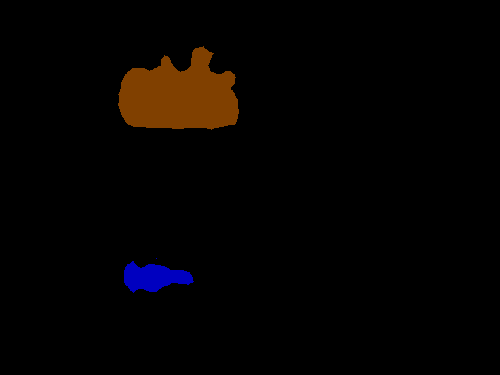}
	\end{subfigure}
	\begin{subfigure}{0.24\linewidth}
		\centering
    	\includegraphics[width=0.9\linewidth]{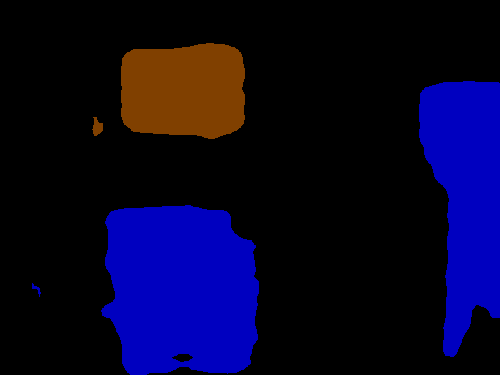}
	\end{subfigure}\\
	\centering
    \begin{subfigure}{0.24\linewidth}
		\centering
    	\includegraphics[width=0.9\linewidth]{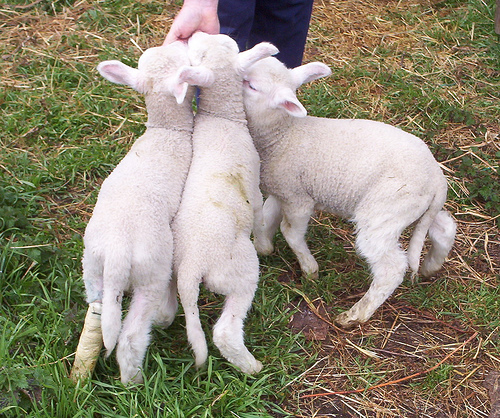}
	\end{subfigure}
	    \centering
    \begin{subfigure}{0.24\linewidth}
		\centering
    	\includegraphics[width=0.9\linewidth]{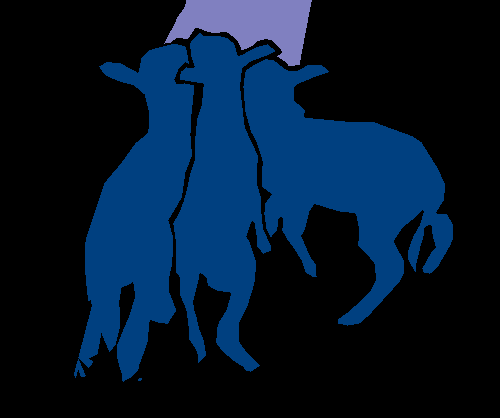}
	\end{subfigure}
		\centering
    \begin{subfigure}{0.24\linewidth}
		\centering
    	\includegraphics[width=0.9\linewidth]{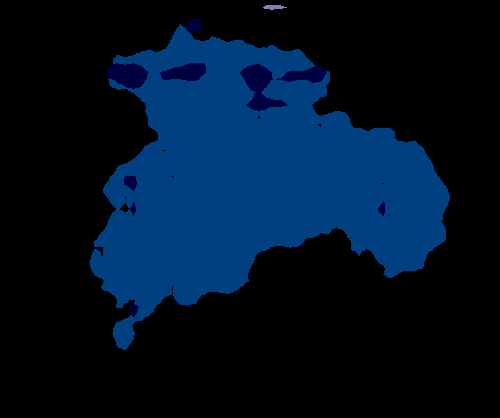}
	\end{subfigure}
	\begin{subfigure}{0.24\linewidth}
		\centering
    	\includegraphics[width=0.9\linewidth]{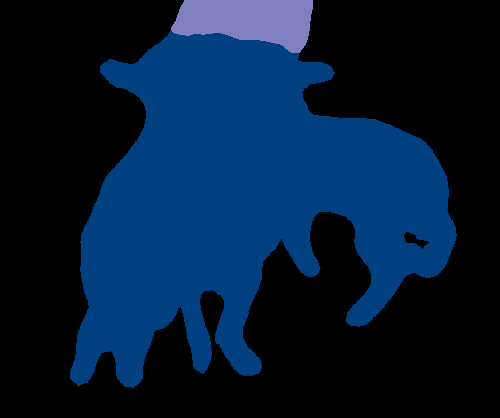}
	\end{subfigure}\\
	\centering
    \begin{subfigure}{0.24\linewidth}
		\centering
    	\includegraphics[width=0.9\linewidth]{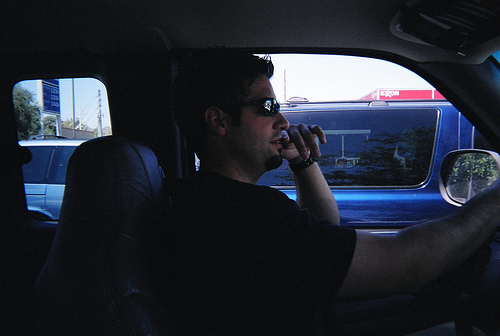}
	\end{subfigure}
	    \centering
    \begin{subfigure}{0.24\linewidth}
		\centering
    	\includegraphics[width=0.9\linewidth]{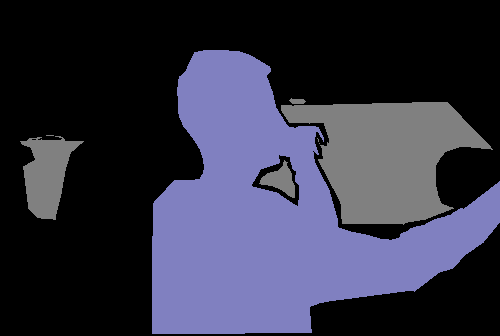}
	\end{subfigure}
		\centering
    \begin{subfigure}{0.24\linewidth}
		\centering
    	\includegraphics[width=0.9\linewidth]{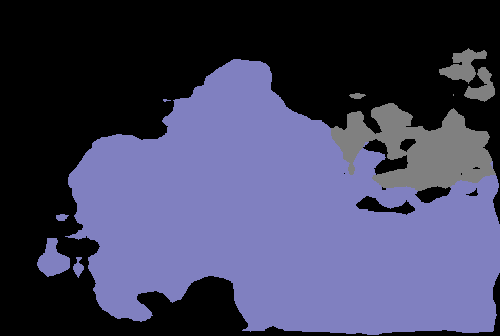}
	\end{subfigure}
	\begin{subfigure}{0.24\linewidth}
		\centering
    	\includegraphics[width=0.9\linewidth]{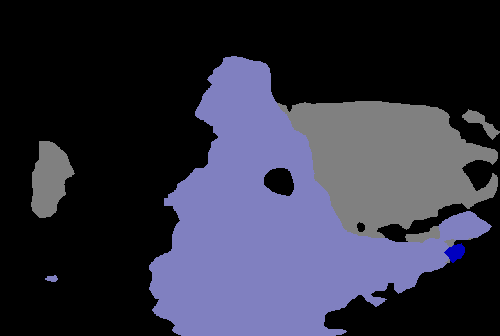}
	\end{subfigure}\\
	\centering
    \begin{subfigure}{0.24\linewidth}
		\centering
    	\includegraphics[width=0.9\linewidth]{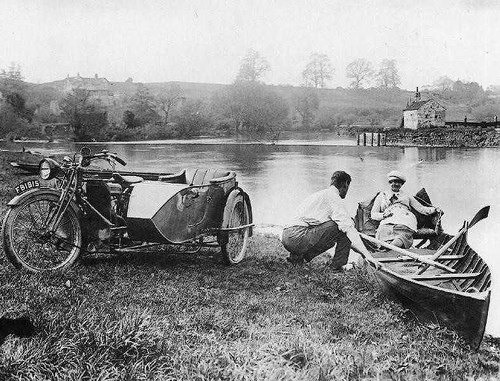}
    	\caption{Image}
	\end{subfigure}
	    \centering
    \begin{subfigure}{0.24\linewidth}
		\centering
    	\includegraphics[width=0.9\linewidth]{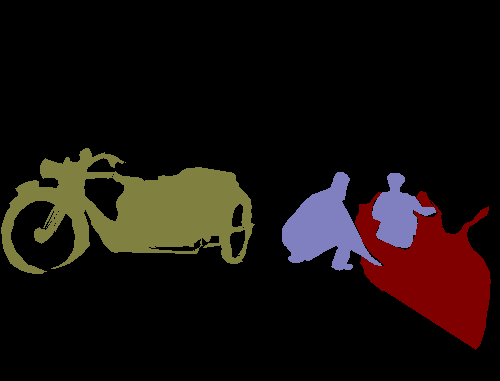}
    	\caption{Ground Truth}
	\end{subfigure}
		\centering
    \begin{subfigure}{0.24\linewidth}
		\centering
    	\includegraphics[width=0.9\linewidth]{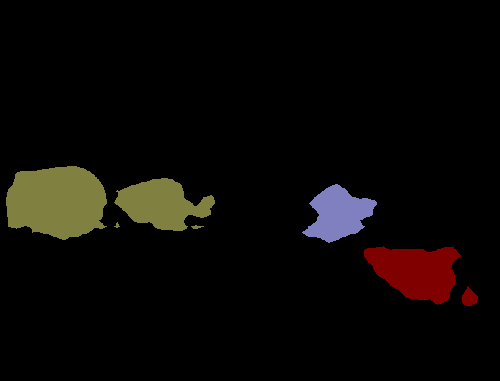}
    	\caption{Random Selection Strategy}
	\end{subfigure}
	\begin{subfigure}{0.24\linewidth}
		\centering
    	\includegraphics[width=0.9\linewidth]{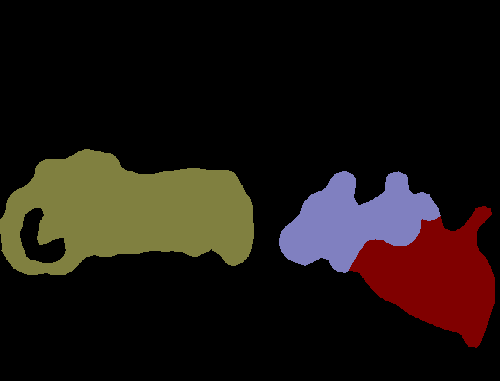}
    \caption{Ours}
	\end{subfigure}
	\caption{The segmentation visualization comparison results comparison between our method with random selection strategy.}
	\label{visual}
\end{figure*}

In Fig.\ref{visual}, we present the segmentation results on the Pascal-VOC 2012 validation set using the model trained in the CSS task. We compare our method with the replay approach using the randomly selected samples. Thanks to the proposed mechanism that automatically learns an optimal policy and uses it to select and enhance the most adequate samples, our method can be more effective to alleviate the catastrophic forgetting problem in CSS, thus achieving the better results. 

\end{document}